# Deep Transfer Learning & Beyond: Transformer Language Models in Information Systems Research

Deep Transfer Learning & Beyond

Transformer Language Models in Information Systems Research

## Ross Gruetzemacher

Wichita State University, W. Frank Barton School of Business, ross.gruetzemacher@wichita.edu

## David Paradice

Auburn University, Harbert College of Business, dparadice@auburn.edu

AI is widely thought to be poised to transform business, yet current perceptions of the scope of this transformation may be myopic. Recent progress in natural language processing involving transformer language models (TLMs) offers a potential avenue for AI-driven business and societal transformation that is beyond the scope of what most currently foresee. We review this recent progress as well as recent literature utilizing text mining in top IS journals to develop an outline for how future IS research can benefit from these new techniques. Our review of existing IS literature reveals that suboptimal text mining techniques are prevalent and that the more advanced TLMs could be applied to enhance and increase IS research involving text data, and to enable new IS research topics, thus creating more value for the research community. This is possible because these techniques make it easier to develop very powerful custom systems and their performance is superior to existing methods for a wide range of tasks and applications. Further, multilingual language models make possible higher quality text analytics for research in multiple languages. We also identify new avenues for IS research, like language user interfaces, that may offer even greater potential for future IS research.

## 1 INTRODUCTION

There is tremendous hype about artificial intelligence (AI) and its potential to transform business. However, many organizations have struggled to see real benefits to their bottom lines due to AI initiatives [Fountaine 2019]. While Fountaine et al. are correct to suggest that organizations need to change their culture to reap the benefits of AI, it is also true that many of the benefits of AI have yet to be realized because the technology is still in its nascency and research progress continues at a rapid pace. There is no apparent reason to suspect this progress to slow, either, and leading organizations in business consulting, economics and policy all foresee AI-driven transformative change in business on the horizon.

Rapid progress in the use of deep learning – the AI technique driving current progress – for image processing and speech recognition in the early-to-mid-2010s was impressive, and progress in deep reinforcement learning has drawn a lot of media attention by demonstrating superhuman performance in a number of games [Silver 2017; LeCun 2015]. Yet, it is debatable as to whether the perceived progress is living up to the hype in practice. While deep learning certainly has valuable applications in business operations and business analytics [Kraus 2020], it has not yet led to significant productivity gains [Brynjolfsson 2020].

However, there is reason to think that recent progress in the use of pretrained language models, which emerged in 2018, may be different. Anya Belz, in the opening keynote of the 2019 International Conference on Natural Language Generation, only days after the release of the most powerful generative language model to date, T5 [Raffel 2020], openly asked "Did T5 just solve general natural language generation?" [Belz 2019]. This question was not made in jest, rather, it was delivered with a sense of dismay; progress truly is being made at a rate which many who have worked on the problem for a long time find unnerving. T5 is no longer the most powerful generative language model, and its successor, GPT-3 [Brown 2020], may have improved even further beyond T5 than it had improved beyond its predecessors.

The three primary reasons to believe that this recent progress is different are not evidenced by the nature of the progress alone but in large part by the nature of how these techniques naturally fit into organizations' operations. *First*, organizations create and collect large amounts of unstructured data. This data is widely thought to contain information that, if harnessed, could be very valuable. For this reason, even moderately effective text mining techniques are already able to deliver tremendous value to organizations in numerous domains ranging from policy [Ngai 2016] to finance [Kraus 2017] to biomedical engineering [Gonzalez 2015]. *Second*, this new generation of pretrained language models harnesses the enormous potential of unsupervised and semi-supervised learning [Collobert 2008]. This means that these models can be initially trained in an unsupervised fashion on very large corpora and then later they can be fine-tuned (*i.e.*, **deep transfer learning**) on an organization's labeled, task-specific data so that they outperform existing text mining techniques for a variety of tasks specific to the organization's needs. *Third*, progress in this area is showing no signs of slowing down, and more advanced capabilities from increasingly powerful systems may continue for some time [Kaplan 2020]. Examples include chatbots' capabilities which are likely to bring long anticipated **language user interfaces** (LUIs) [Brennan 1991] to a wide variety of human-computer interactions, and **few-shot learning** capabilities that can reduce the training and skills required for using language models while creating the possibility of truly novel applications.

This paper is intended for researchers and practitioners who are interested in any type of business research that may benefit from analysis of large amounts of unstructured text data (*e.g.*, emails, reviews, social media posts) as well as those interested in applications of LUIs, both practically and in information systems (IS) research. This study makes several contributions:

1. It identifies and reviews the state-of-the-art literature for a powerful application of deep learning that has not yet been effectively incorporated into the toolbox of IS researchers.
2. It conducts a literature review of existing work in leading IS journals using text mining, clearly identifying limitations of existing work and the benefits of using the new tools.

3. It proposes concrete research ideas that go beyond simply improving existing work to offer new directions for researchers and practitioners to explore.

No text mining or NLP experience is necessary for readers of this article[1], but we do assume that readers are familiar with the concepts of neural networks and deep learning. In the remainder of the paper we first survey the recent progress that has led to these powerful new tools. We next survey extant applications of text mining for business analytics and IS research. We then consider recent applications of these new tools within a discussion of their implications for both research and practice. We follow the discussion by summarizing its salient elements, including the most promising avenues for future work, and finally we leave concluding remarks.

## 2  RECENT PROGRESS IN NEURAL LANGUAGE MODELS

The subfield of machine learning known as computational linguistics or natural language processing (NLP) has been one of the primary focuses for AI researchers since the beginning of the study of AI: the first conference on machine translation preceded even the 1956 Dartmouth workshop, thought of as a seminal event for the birth of the field, and the necessity of NLP for AI was clear as early as Turing's proposed test for intelligence (a.k.a. the Turing test) [Turing 1950]. The early years of NLP research (*i.e.*, 1960-1985) centered on what is known as the rationalist approach. Statistical NLP, which takes an empiricist's approach, did not become the dominant school of thought until the 1990s [Manning 1999]. Statistical NLP assumes that a large degree of latent semantic knowledge resides in text corpora, and, in order to encode this knowledge, numerical representations of language are necessary [Smith 2020]. Such numerical representations of words are called **word representations** (a.k.a. word vectors or word embeddings), and they are a fundamental building block of statistical NLP.

Originally, words were encoded simply by assigning an integer to each unique token. However, integers are poor word representations because they do not allow semantic information to be shared across words with similar properties [Smith 2020]. Multidimensional vectors, on the other hand, can contain continuous values for each dimension, and these dimensions can be thought of as semantic features of the word being represented. For example, if we assign a dimension of a word vector to be associated with weight (in grams), *feather* might have a value of 0.001, *penny* might have the value of 2.5 and *car* might have a value of 1,250,000.0, but adjectives like *green* and *chilly* would have values of zero. Creating representations of words in this way is known as feature engineering, but it is not practical for most corpora, and surely not for an entire language. In most cases, learning word representations is far more useful. Early semantic representations utilized frequency-based methods like singular value decomposition of the co-occurrence matrix. Such approaches power many widely used text mining techniques like latent Dirichlet allocation (LDA) [Blei 2003]. These techniques comprise one family of word representations called global matrix factorization models [Pennington 2014].

Using word representations to encode the semantics of words in a language, **statistical language models** (a.k.a. probabilistic language models or simply language models) can be created to model the probability of word occurrence in sentences. Specifically, language models are probability distributions of sequences of words that are useful for problems that require the prediction of the next word in a sequence given the previous words. n-gram models are a very simple form of language model that are commonly used in text mining, and such simple models have long been used for a variety of other tasks including spell check, machine translation and speech and handwriting recognition [Manning 1999].

---

[1] We do not go into technical details here as our purpose is to inform readers about the possible applications of these techniques, both now and in the future, relative to existing text mining techniques.

## 2.1 Neural Word Representations

Neural language models comprised of distributed word representations were first proposed as a solution for the curse of dimensionality [Bengio 2003]. Collobert and Weston [2008] then demonstrated the value of using deep learning for learning distributed representations of words from large unlabeled corpora, then transferred the learnt knowledge to multiple tasks learned simultaneously through further training (*i.e.*, fine-tuning) on labeled datasets (i.e., deep transfer learning). Early last decade, Collobert et al. [2011] described the first pretrained neural word representations that were able to achieve strong performance on major NLP tasks.

In the time since these early studies, distributed word representations have become widely preferred over alternate representations. One reason feature engineering is not practical for NLP is due to the challenge of identifying all of the relevant features of words that would need to be represented in order to capture the entire semantics of a corpus or language. However, representation learning generates a latent feature space where features are not constrained by the need to map directly to human concepts in natural language, which makes it much easier to capture the rich semantics of a language with a limited number of features (*e.g.*, 100 to 300).

The first strong neural language model to learn practically useful word representations in this manner was word2vec [Mikolov 2013b]. For training, Mikolov et al. proposed two different architectures: one for predicting the current word based on context (*i.e.*, continuous bag-of-words or CBOW) and another for predicting the surrounding words given the current word (*i.e.*, continuous skip-gram). The former was better suited for small corpora while the latter was better suited for scaling to large corpora. The new techniques proposed by Mikolov et al. were able to generate rich word representations that captured fine-grained semantic and syntactic regularities better than previous models. This led to the widespread use of word representations in NLP.

The ability of these neural word representations to explicitly encode numerous linguistic regularities and patterns exhibited some very interesting characteristics: the relationships between two words could be represented as linear translations and that simple vector operations could be used to evaluate the concepts of semantic and syntactic similarity between words [Mikolov 2013a]. For example, the vector operation $\overrightarrow{Madrid} - \overrightarrow{Spain} + \overrightarrow{France}$ was closer to $\overrightarrow{Paris}$ than any other word. Even vector addition alone had valuable results: in the latent feature space $\overrightarrow{Germany} + \overrightarrow{captial}$ was close to $\overrightarrow{Berlin}$ and $\overrightarrow{Russia} + \overrightarrow{river}$ was close to $\overrightarrow{Volga\ River}$.

word2vec [Mikolov 2013b] was the first in a new family of word representation models that are very useful for analytics because it enabled building custom models from rich, learnt semantic features. However, it was not without limitations. For example, it did not leverage document level information during training. Alternately, global matrix factorization models were able to leverage document level statistical information but were unable to perform well on the analogical evaluation in which word2vec excelled (*i.e.*, vectors' semantic and syntactic similarity). Pennington et al. [2014] attempted to these issues with the GloVe (global vectors for word representation), which was able to make efficient use of document level statistics like global matrix factorization methods while also generating representations with a meaningful vector space substructure like that of word2vec. GloVe outperforms word2vec on some benchmarks, but both techniques are still widely used.

We often refer to word representations like word2vec and GloVe [Mikolov 2013b; Pennington 2014] as pretrained word representations because these word representations are openly available for public download having already been pretrained on large text corpora. When used in this fashion they can be very powerful because the Web-based corpora that they have been trained on are often too large for training by independent researchers with limited computational resources. Even for those with the requisite computing power, these pretrained representations save time from tuning hyperparameters and cleaning corpora. However, it is also common practice to train representations on smaller, domain-specific corpora which require fewer computational resources. This can be worth the time spent tuning hyperparameters due to the improvements that can be obtained by using domain-specific word representations.

## 2.2 Neural Language Models

Traditional deep neural networks[2] are not well suited for language processing because NLP tasks often require mapping from vectors of different lengths. For example, a language model designed to predict the next word in a sentence must operate on an input of one word as well as an input of 20 words in order to predict the next word. Recurrent neural networks (RNNs) are neural networks with feedback connections that are suitable for machine learning tasks requiring such sequence-to-sequence mapping [Murphy 2012]. Their suitability for these tasks is due to the fact that, unlike normal neural networks, they are able to map from vectors of varying lengths to other vectors of varying lengths. One of the challenges that RNNs face in NLP is known as the vanishing gradient, which can cause training to fail, but there are techniques that can be used to mitigate this problem. Long short-term memory (LSTM) [Hochreiter 1997] models are a form of RNN that use a gating mechanism to address this problem, and they have long been used for a large number of NLP applications. An LSTM gating mechanism controls the flow of information to hidden neurons, which, for sequence-dependent data (e.g., sentences), can encode the meaning of a sequence while remembering (or forgetting) the most (or least) salient elements. Other approaches can also be applied to RNNs to counteract the vanishing gradient problem [Mikolov 2014], such as the gated recurrent unit [Chung 2014], but none have been as effective as the LSTM. LSTMs work very well for a variety of challenging tasks; however, LSTMs typically rely on supervised learning and require a unique labeled training dataset for each task. They are well-suited for a variety of NLP tasks ranging from classification to translation to text generation and were the dominant NLP technique prior to the development of attention and the transformer architecture[3].

All of the word representations described thus far are static word representations, and they all have one major limitation: they attempt to represent words in all possible contexts with a single vector. However, words have different meanings in different contexts and thus are not always best represented in a static manner. **Contextual word representations** (CWRs) offer a solution for this based on the premise that if each word is going to have a unique representation then each vector should be dependent on a separate context vector representing the sequence of nearby words. CWRs were popularized by Peters et al. [2018] with the ELMo language model. ELMo was significant in that it demonstrated state-of-the-art (SOTA) performance on not just one NLP task but six, suggesting that performance gains from this approach were likely for a wide variety of NLP tasks. It also ushered in a class of pretrained language models with rich word representations embedded in the weights. *Because these new language models are both language models and (able to generate) rich word representations, we refer to them simply as language models.*

---

[2] Deep learning is a form of representation learning: a type of machine learning involving learning of representations or features in data. Mathematically, it can be thought of as a technique for learning a function to map from the input data to the output data.

[3] While LSTMs are thought to have been the dominant technique prior to the transformer, convolutional neural networks were (and are) still very capable and even preferable for many applications (e.g., classification). Recent work [Tay 2021] suggests that convolutional neural networks may still have significant capabilities despite the recent prevalence of transformer-based language models.

### 2.3 Deep Transfer Learning

Transfer learning has long been a topic of interest for machine learning researchers. In fact, it predates deep learning and is a machine learning problem unto itself. However, deep transfer learning is very powerful and has become a topic of interest, being used for tasks from image processing to NLP.

Generally speaking, transfer learning involves the transfer of knowledge learned from one learning task to improve results or speed up training for another task. It effectively removes the need to train a model from scratch by enabling specialized training for a new task via the fine-tuning of an existing, pretrained model. **Deep transfer learning** (DTL) refers to this use of deep learning for pretraining models on large amounts of data, either from labeled data for supervised pretraining or from unlabeled data for unsupervised pretraining. These pretrained models can then be fine-tuned on task-specific datasets to transfer the knowledge learnt from the more general, original training datasets to the domain-specific applications and use cases.

**Pretraining**, as it is most commonly used for learning word representations, is a specific form of unsupervised learning known as self-supervised learning. Self-supervised learning does not require explicit labels for data as supervised learning techniques do, rather, implicit supervisory signals from the data are autonomously extracted and used during pretraining. In the case of NLP, these signals come from the sequence of the words, e.g., a model can be trained by masking a single word and training the model to predict that word given the surrounding words. Pretraining is critical to DTL model performance, and there are several key aspects that are important to understand. Typically, pretraining is performed using very large corpora, and corpora selection or curation can have a significant impact on model performance and end tasks. Also, the selection of a pretraining objective and the approach for self-supervision can have significant impacts on model performance and end tasks. The relevance of these elements will be explained further in the following sections.

**Fine-tuning** is a critical component of DTL, too. It refers specifically to the further training of a pretrained model on a smaller, labeled dataset. For NLP it refers specifically to the process of leveraging the vast semantic knowledge contained in large, pretrained models for application to domain-specific tasks involving small domain-specific datasets. It is valuable because it enables the simple development of custom, SOTA NLP systems for a great variety of tasks with relatively small, labeled datasets and with significantly less effort than previous techniques (*e.g.,* LSTMs). Oftentimes with the latest language models SOTA performance for a task can be attained by simply fine-tuning on task-specific datasets [Howard 2018].

In the context of language models, transfer learning is considered to have four steps: pretraining, further pretraining, pre-finetuning and fine-tuning. The first and last steps have been explained, but the steps in-between can be useful for significantly improving performance when using DTL. Further pretraining involves pretraining an already pretrained model on an alternate dataset, commonly one which is smaller and either domain-specific or task-specific. While more computationally expensive than fine-tuning, further pretraining is still much less resource intensive as pretraining the model from scratch and can be worth the cost when end task performance is critical. Additionally, further pretraining differs from initial pretraining in that further pretraining does not involve self-supervised signals and is not full self-supervised learning like initial pretraining.

Like further pretraining, pre-finetuning is performed on an already pretrained model in order to further refine representations prior to end-task fine-tuning. It involves the use of a broad supervised dataset for multitask training in order to encourage learning representations that will generalize better to a variety of downstream tasks [Aghajanyan 2021]. Less computationally expensive than further pretraining, pre-finetuning can be used to improve performance when end-task effectiveness is critical or to improve zero-shot performance [Wei 2021].

## 2.4 Transformer Language Models

Because the study focuses on transformers and transformer-based language models, we split this section into two subsections: a high-level overview of recent progress, consistent with the overall narrative, and a discussion of more technical details about the transformer and models based on it that are relevant to IS researchers.

### 2.4.1 Overview

In late 2017 the transformer architecture was first proposed by Vaswani et al. [2017]. At the time LSTMs were the prevailing paradigm in NLP, but they did not work terribly well or reliably for very long sequences or transfer learning. The transformer presented a novel way to incorporate an attention mechanism in deep feedforward networks that allowed it to capture long range sequence dependencies like the LSTM, but with a larger context window for longer sequences. The transformer was also easily parallelizable and highly scalable. Due to this, transformers have been trained using unprecedently large corpora [Raffel 2020; Brown 2020]. We distinguish language models using the transformer as **transformer language models** (TLMs) because they perform remarkably better than LSTM-based models and they scale very well [Kaplan 2020].

There are a large number of TLMs, but here we will initially focus on three of the most significant with respect to practicality, novelty and improvement upon previous models[4]. TLMs first emerged over summer of 2018 [Radford 2018], but the most powerful early model was BERT (bidirectional encoder representations from transformers[5]) [Devlin 2018], which was demonstrated in late 2018. In the time since it has become the most widely used TLM[6], and is able to achieve SOTA performance on a wide number of tasks due to its versatility.

The next major model advance was the text-to-text transfer transformer (T5) [Raffel 2020], which was developed specifically for transfer learning and is designed to operate solely through text generation by framing all text-based language problems as text-to-text tasks[7]. We refer to language models like this that operate solely through text generation as generative language models. In contrast to models like BERT where fine-tuning involves adding a fully connected layer and output neurons, which means that separate models are necessary for multiple tasks, T5 is intended to be fine-tuned on multiple tasks by default. By design T5 can be trained on multiple tasks simultaneously, in the fashion proposed by Collobert and Weston [2008] over a decade earlier.

The final language model we mention here is also a generative language model: the generative pretrained transformer 3 (GPT-3) [Brown 2020]. GPT-3 is an OpenAI TLM which uses the same architecture as its predecessor, GPT-2 [Radford 2019]. What makes GPT-3 unique is its scale: GPT-3 was scaled to a model size, measured by number of parameters, an order of magnitude larger than any previous model and was pretrained on the largest dataset to date. This required extreme investments in computational resources and distributed computing infrastructure, but led to surprising improvements on zero-, one- and few-shot learning tasks.

---

[4] TLMs are trained in a variety of model sizes (number of parameters). For this survey we only consider the largest of each TLM.

[5] BERT has two notable characteristics: it is trained bidirectionally and it is a masked language model. Instead of being trained for predicting the next word in a sentence it is trained to predict missing words in a sentence. It does so by masking (*i.e.*, masked language model) 15% of the words and training bidirectionally to predict the missing words as well as the next sentence.

[6] There are now a large number of variants of BERT, either architectural variations or models that were pretrained on a domain-specific dataset. BERT is so popular that a survey [Xia 2020] on the different variants and how to pick the best one for different types of problems was published recently at one of the premier NLP conferences. (This survey is a valuable resource).

[7] T5 was the result of a large-scale study by Google on the limits of transfer learning from transformer language models and it is not like previous models because it operates as a text-to-text language model, meaning that it both receives text as an input and produces text as an output. Most NLP tasks can be formulated in this manner, and this enables T5 to train a single model to perform multiple tasks during inference by appending a label associated with each unique task to the beginning of the input text.

**Few-shot learning** refers to the ability of a system to be able to learn without the need for even modestly sized datasets typically used for fine-tuning. For example, being a generative language model, the model could be trained by providing ($k$) questions as input and the ($k$) correct answers as the training targets (*e.g.*, for one-shot learning $k$ is one). For two-digit addition problems GPT-3 achieves 99.6% accuracy with only one example, and with no examples – *i.e.*, zero-shot learning – GPT-3 still achieves 76.9% accuracy. Few-shot learning would involve $k$ training prompts and targets, with a limit set by the fixed context window of 2048 tokens. Thus, if the model does not perform well on tasks with zero- or one-shot learning, more examples can be used to improve performance. Further work from OpenAI suggests language model performance will continue to scale with computational resources and dataset size, with no plateau in sight [Kaplan 2020].

### 2.4.2 Under the Hood

The original transformer proposed by Vaswani et al. [2017] includes both encoder and decoder components. The encoder encodes the input sequence into a high dimensional feature space, and the decoder converts high dimensional representations back into words. This is known as a sequence-to-sequence (seq-to-seq) model, and, as such, it is naturally well suited for tasks like machine translation. However, the key contribution of the paper was not the encoder-decoder element, but rather that, unlike the LSTM, transformers do not use recurrence or require any sequential computation. Thus, transformers are not subject to the vanishing gradient problem[8]. Attention mechanisms predate the transformer [Xu 2016], but the transformer made practical use of attention in a novel and powerful way that enabled SOTA results on a major machine translation benchmark.

Pretraining is the most critical element of a TLM as this is what distinguishes different TLMs. However, TLMs typically fall into one of three categories with respect to their pretraining: autoencoding, autoregressive or seq-to-seq [Wolf 2019]. BERT is an example of an autoencoding TLM whereas GPT-3 is an example of an autoregressive TLM, and the original transformer is an example of a seq-to-seq TLM. The distinction between models is determined by the pretraining scheme. For example, BERT encodes documents bidirectionally and replaces random tokens with masks, then is trained to predict masked tokens as well as the next sentence. This contrasts with GPT-3, where tokens are predicted autoregressively with a left-to-right decoder. Autoencoding models perform best at discriminative tasks (e.g., classification, regression; tasks where BERT excels) while autoregressive models perform best at generative tasks (e.g., summarization, dialogue; where GPT-3 excels).

BART is an example of a seq-to-seq model that attempts to bridge the divide between autoencoding and autoregressive models [Lewis 2020a]. It used a variety of denoising schemes for pretraining its denoising autoencoder. The results suggested that new pretraining schemes could lead to strong performance on generative tasks without sacrificing performance on discriminative tasks, and that different approaches to corrupting documents[9] during pretraining may be better suited for specific downstream tasks. Another seq-to-seq TLM based on a new type of denoising autoencoder, MARGE [Lewis 2020b], utilizes an alternative self-supervision technique to the dominant token masking paradigm; similar documents from other langauges are used to assist in reconstruction of the input document. MARGE performs strongly on a wider range of tasks in many languages – both discriminative and generative – than previous models.

---

[8] While the LSTM uses a gating mechanism to mitigate the problem of the vanishing gradient, this does not overcome it completely, and the vanishing gradient still limits the context window of the LSTM. By not using recurrence at all, the transformer avoids this problem which results in a larger context window and the transformer's most significant improvement over the LSTM.

[9] By corrupting we are referring to pretraining approaches such as masking 15% of input tokens during training, as with BERT. A variety of corruption approaches are explored with BART, and different approaches could have impacts on downstream tasks with fine-tuning.

### 2.5 Natural Language Understanding

Natural language understanding (NLU) is typically thought to be a more general, longer-term goal for NLP researchers. We mention it here because significant effort has been made to develop measures to quantify progress in this domain, and these measures demonstrate the recent progress of TLMs. In April of 2018, researchers from leading institutions in business and academia realized the need for a new means of assessing progress and developed the General Language Understanding Evaluation (GLUE) benchmark [Wang 2018], which was intended to be a benchmark for measuring progress toward NLU. Just over a year later, in June of 2018, Microsoft had surpassed the human baseline for GLUE [Liu 2019a]. However, this was anticipated, and a more difficult SuperGLUE benchmark was released [Wang 2019a].

BERT was used as the initial baseline for the SuperGLUE benchmark achieving a score of 69.0, well below the human baseline of 89.8, but less than three months later a team from Facebook AI Research had demonstrated a robustly optimized version of BERT (RoBERTa[10]) that was able to achieve a SuperGLUE score of 84.6[11] [Liu 2019c]. This striking progress led to speculation that, like progress in other domains such as self-driving vehicles, the first 95% of the task was less difficult than previously perceived, but that the last 5% would become exponentially more challenging. However, less than three months later, and to the dismay in the natural language generation community [Belz 2019], T5 was released demonstrating a score of 88.9[12] on SuperGLUE – within a point of the human performance baseline [Raffel 2020].

Following the release of T5 [Raffel 2020], progress appeared to slow for six months, which seemed to suggest that early intuition about the last 5% becoming more difficult may be valid. However, this lag was again shown to be unfounded by GPT-3 [Brown 2020]. *GPT-3 impressed for many reasons, but its performance might be best summarized by considering that it achieved a SuperGLUE score of 71.8, over 4% higher than BERT, simply by using few-shot learning (k = 32).* As impressive as this is, it is also important to consider that the variance among scores for the different tasks comprising the aggregate measure was dramatically higher for GPT-3. For example, BERT outperformed GPT-3 by over 45% on one complex linguistic task but GPT-3 outperformed BERT by over 30% on a widely used causal reasoning benchmark.

T5, with the use of an Unsupervised Data Generation (UDG) procedure [Wang 2021b], has since surpassed human-level performance on SuperGLUE. Other models have also exceeded human-level performance on SuperGLUE, including a decoding-enhanced BERT with disentangled attention (DeBERTa) from Microsoft [He 2021], which builds on RoBERTa with disentangled attention and enhanced mask decoder training. Another multilingual model, ERNIE 3.0 [Sun 2021], from Baidu, has even surpassed the SuperGLUE performance of T5. To achieve this feat, ERNIE 3.0 fuses an auto-regressive network and an auto-encoding network for training on both text data and a large-scale knowledge graph.

One research topic that clearly falls under the purview of NLU is chatbots. Progress in this area has also seen great improvements since 2018, with the most significant progress occurring recently. At the beginning of 2020 Adiwardana et al. [2020] demonstrated Meena, an open domain chatbot that was thought to have more "human-like" conversation as measured by a novel metric that required evaluation of responses through crowdsourcing. Only a few months later, Facebook AI released Blender Bot open source [Roller 2020a], which was significant as Blender Bot outperformed Meena on human judged evaluations. Because Blender Bot is a TLM-based chatbot, it can be fine-tuned for domain-specific tasks, offering new opportunities for IS researchers.

---

[10] RoBERTa was simply a replication study of BERT that explored the significance of different hyperparameter choices and training dataset size. They found that doing away with the next sentence prediction in pretraining, and some other training modifications, greatly improved the performance of BERT.

[11] RoBERTa should be used instead of BERT when possible due to the easy performance gains from pretraining.

[12] In the time since, T5 has been retrained to score even higher on SuperGLUE with an 89.4.

## 2.6 Ongoing Research and Future Directions

The review thus far brings readers up-to-speed with respect to where the SOTA is for TLMs and NLP. This section discusses some focal areas of current research where substantial progress is being made that stands to dramatically improve future capabilities beyond what might be anticipated from the research discussed prior.

One of the most pressing limitations of TLMs is that they are terribly inefficient; the largest models can only be fine-tuned on the cloud, which can be costly for researchers, and such models are not possible to use on edge or mobile devices. However, important work has been conducted to address this. For example, DistilBERT [Sanh 2019] is 60% faster and 40% smaller than BERT but still retains 74% of BERT's performance above the GLUE baseline. Other models focus on efficiency for specific tasks, such as TopicBERT, which is intended for document classification and achieves a 40% speedup while retaining 99.9% of BERT's performance on five tasks [Chaudhary 2020]. Further, new work from Schick and Schütze [2020] demonstrates strong few-shot learning performance on SuperGLUE using ALBERT (a lite BERT) [Lan 2019], one of the most efficient TLMs.

When Vaswani et al. [2017] first demonstrated the transformer it SOTA results on a major machine translation benchmark. TLMs work well for translation because they can represent the semantics of multiple languages in a shared, high-dimensional latent space. However, there are other applications of multilingual TLMs, and recent work has begun focusing on massively multilingual TLMs trained on 8 languages or more, for which new multi-domain and multi-task benchmarks have been developed [Hu 2020; Liang 2020]. The most widely used multilingual TLM is XLM-R, which is pretrained using text from 100 languages [Conneau 2020]. Optimizing these models for multiple tasks across multiple languages is one of the biggest challenges for multilingual TLMs, but the latest work has demonstrated substantial progress in this direction [Wang 2021a].

While revolutionary, the transformer is far from perfect. One significant limitation is the fixed length context window, which, while far greater than the LSTM, could be even more useful if extended. Early work addressing this, the Transformer-XL [Dai 2019], is used in XLNet [Yang 2019b], which demonstrates superior performance on general sentiment analysis. Other recent work has demonstrated a TLM with a context window of up to one million words [Kitaev 2020]. Another major limitation of transformers is the poor scalability of the self-attention mechanism. To address this, Zaheer et al. [2020] have proposed BigBird, a sparse attention mechanism that drastically improves upon the original transformer's attention mechanism, which could have practical implications for tasks such as longer document summarization and question answering. Choromanski et al. [2021] have gone further proposing another sparse attention mechanism that demonstrates generalized attention which may lead to even greater improvements once used for training large TLMs.

While the transformer has been exploited extensively as the current architecture of choice for the pretrain/fine-tune paradigm, recent work has shown that convolutional neural networks show promise in this area as well [Tay 2021]. This work suggests that architectural progress should be differentiated from progress in pretraining, and that convolutional neural networks outperform TLMs in some cases. Thus, further work is merited to explore alternative architectures to the transformer[13] within the pretrain/fine-tune paradigm.

---

[13] This survey focuses on TLMs because only TLMs alone have demonstrated tremendous progress in NLP – at an equivalent level to AlexNet [Krizhevsky 2012] in computer vision last decade – however, if alternative architectures are as successful, the directions for future IS research suggested in later sections would still apply.

Finally, while much progress has been made in the domain of NLU, it is clear that there are limitations to distributional semantics (*i.e.*, purely probabilistic models of semantic similarity). Grounded semantics refers to the grounding of semantic concepts to knowledge learnt from other forms of data (*e.g.,* images, video, simulation), and is thought to be the next step toward NLU. Early work in this direction has focused on improving "commonsense reasoning" through grounding CWRs via training a multimodal model on a question answering dataset about images from movie scenes [Zellers 2019]. Recent work from Tan and Bansal [2020] builds on their previous work on cross-modal TLMs (LXMERT) [Tan 2019] by exploring the possibility of a visually-supervised TLM through a process they call "vokenization."

Visual grounding may produce impressive results at present, but effective communication relies on a shared understanding of the world, one which is learnt from experience. Drawing from this notion, Bisk et al. [2020] have proposed five levels of *World Scope*: 1) corpus (the past), 2) the Web (most of current NLP), 3) perception (multimodal NLP), 4) embodiment (situated action taking), and 5) the social world. This new perspective is exciting because it situates existing work and identifies the next steps forward toward the shared understanding of the world necessary for truly successful linguistic communication through language grounding.

## 2.7 Summary

After the preceding overview of statistical NLP up to the current state of the field, we want to highlight three distinct periods through which recent progress can be better understood. The first period began with the inception of the field and lasted until 2013. This period was characterized in large part by hand crafted features and to some degree the emergence of statistical NLP. The bulk of text mining techniques used in practice today originated in this period. The second period began in 2013 with the word2vec model [Mikolov 2013b]. This ushered in a new, data-driven period characterized by neural word representations and neural language models. The most recent period began in 2018 and involves the topics which are the focus of this study.

 This current era of NLP is defined by three significant components: 1) CWRs, 2) DTL/fine-tuning and 3) TLMs. Combined, these elements are transforming the study of NLP because they enable leveraging unsupervised learning for a large number of tasks and applications. Fundamentally, this means that there is no limit to the amount of data that can be used for training, which brings new meaning to the phrase "big data". Further, gains from continued scaling of language models are not expected to plateau anytime soon [Kaplan 2020]. Such continued progress combined with grounded semantics advances may usher in truly transformative language processing, and it is essential for IS researchers to be familiar with progress in this domain.

## 3  TEXT MINING IN INFORMATION SYSTEMS RESEARCH

As this paper explores the future applications of TLMs in IS research, we felt it appropriate to conduct a thorough review of recent work in IS that used text mining or NLP. Due to the widespread application of these techniques, we limited our review to articles published (and preprints accepted for publication) from 2016 to 2020 in the three leading IS journals: Information Systems Research, the Journal of MIS and MIS Quarterly[14]. We chose to focus our review on three terms: "text mining," "sentiment analysis" and "natural language processing." We queried these terms using Google Scholar's advanced search feature on September 10th, 2020. Only work that utilized these methods as part of a model were included in the results – studies that just mentioned the terms or only used them for robustness checks were discarded.

---

[14] Even more articles using text mining appeared in other IS outlets such as Decision Support Systems and the International Conference on Information Systems proceedings. However, only articles from the top three IS journals were selected for inclusion in order to reduce noise because there were so many articles from these other outlets and the articles in the top three journals were deemed to be most representative of rigorous IS research. Moreover, very few articles using text mining appeared in the other basket of eight IS journals.

This process resulted in 55 papers meeting our criteria. These papers are listed in Table I which depicts relevant features of each study. Each paper was carefully reviewed and coded with respect to these relevant features, seven of which characterized the techniques and seven more that characterized other relevant aspects of the studies. Each author independently coded each paper, and for disagreement, a discussion was conducted to arrive at a consensus. For coding we classified each feature with check marks of two shades to represent weak correspondence (grey) and strong correspondence (black). The four results that cited the two original neural word vector models [Mikolov 2013b; Pennington 2014] are shaded, and just one paper [Shin 2020] cited BERT [Devlin 2018], the most widely used TLM, but only as a suggestion for future work. *No papers published or forthcoming in any of these journals at the time of this literature review had utilized CWRs, DTL or TLMs.*

For brevity, we do not discuss the details of the results of the IS literature review reported in Table I, but we will reference different elements of this table in the remainder of the document. In some cases, we reference specific studies from this section to demonstrate how TLMs can be used to improve on this work. However, our review of IS literature is not limited to just these studies, and in the next section we also identify more existing research, from IS and other domains, that did not utilize text mining, but which still stands to benefit from TLMs.

The following section focuses on the most significant ways that TLMs and CWRs could impact IS research. While IS research examples are used in this, more detailed examples are described in Appendix A: Summaries & Recommendations. In this supplementary material, we include a summary of each paper from the IS literature review along with recommendations for how, or how not, the work might benefit from using TLMs or CWRs. We feel that this also can help readers to better understand how these new techniques are poised to impact IS research. This appendix was not included in the main text due to length, but we strongly feel that it is a major contribution of the paper; consequently, we recommend interested readers consider it to be primary content.

While Table I reports the results of the literature review, and supports the content presented in the remainder of the paper, it alone does not provide a valuable synthesis for IS researchers applying text mining and NLP in their research. A critical element of research involving text data is identifying appropriate techniques for different types of data. Table II shows the methods from our literature review that are most used for various types of text analytics employed in IS research. It can be seen that all methods are used for social media data, and that nearly all techniques are also used for reviews. However, for other documents (e.g., government documents, financial documents) and for text data collected from apps, only techniques like topic modeling and feature extraction (or word representation models) have been utilized. However, for more generic internet data that does not involve reviews or social elements, all techniques have been used, except for sentiment analysis.

In the following section we discuss how TLMs, CWRs and DTL can be used for each of the five classes of text mining techniques that we cover in Table II. While we also cover more speculative applications for IS researchers, we feel that simply applying these new techniques in manners consistent with the most common applications from the existing literature offer the most promising opportunities for IS researchers.

## Table I: Analysis of Text Mining Research in IS Journals

| Group | Study | Sentiment Analysis | Topic Clustering | Feature Extraction | Entity Extraction | Text Classification | Emotion Detection | Other | Predictive Model | Reviews | Social Media | Novel Method | Text Mining Focus | Robustness Check | Foreign Language |
|---|---|---|---|---|---|---|---|---|---|---|---|---|---|---|---|
| **Information Systems Research** | Abbasi et al. 2019 | √ |  |  |  |  |  |  | √ |  | √ | √ |  |  |  |
| | Adamopoulos et al. 2018 | √ | √ | √ |  |  | √ |  | √ |  | √ |  |  |  |  |
| | Blohm et al. 2016 |  |  |  |  |  |  | √ |  |  |  |  |  |  |  |
| | Chen, K. et al. 2020 | √ |  |  | √ |  |  | √ | √ |  |  | √ | √ |  |  |
| | Chen, W. et al. 2019 | √ |  |  |  |  |  |  |  | √ |  |  |  |  |  |
| | Chung et al. 2020 | √ |  |  |  |  |  |  |  |  | √ |  |  |  | √ |
| | Heimback & Hinz 2018 | √ |  | √ |  |  |  |  |  |  | √ |  |  |  |  |
| | Hwang et al. 2019 |  | √ |  |  |  |  |  |  |  | √ |  |  |  |  |
| | Khern-am-nui et al. 2018 | √ | √ | √ |  |  |  | √ |  | √ |  |  | √ |  | √ |
| | Lappas et al. 2016 | √ |  |  |  |  |  |  |  | √ |  | √ |  |  |  |
| | Lee et al. 2020 |  | √ |  |  |  |  |  |  |  | √ | √ |  |  |  |
| | Liu, X. et al. 2020 |  | √ | √ |  | √ |  |  | √ |  | √ | √ | √ |  |  |
| | Liu, Y. et al. 2020 |  | √ | √ |  |  |  | √ | √ |  | √ | √ |  |  |  |
| | Mejia et al. 2019 |  |  |  |  | √ |  | √ | √ | √ | √ | √ |  |  |  |
| | Mousavi & Gu 2019 | √ |  |  |  |  |  |  | √ |  | √ |  |  |  |  |
| | Mousavi et al. 2020 | √ |  |  |  |  |  |  | √ |  | √ |  | √ |  |  |
| | Pan et al. 2019 |  |  |  |  |  |  | √ | √ |  | √ | √ |  |  |  |
| | Song et al. 2019 | √ |  | √ |  | √ |  | √ | √ |  | √ |  |  |  | √ |
| | Wang, Q. et al. 2018 |  | √ |  |  |  |  |  | √ |  | √ |  |  |  |  |
| | Yang, Mo. et al. 2018 | √ |  |  |  | √ |  |  |  |  |  |  |  |  |  |
| **Journal of Management Information Systems** | Chen, L. et al. 2019 | √ | √ | √ | √ | √ | √ |  | √ |  | √ | √ | √ |  |  |
| | Dong et al. 2018 | √ | √ | √ |  | √ | √ | √ | √ |  | √ | √ | √ | √ |  |
| | Ghiassi et al. 2016 | √ |  |  |  |  |  |  |  |  | √ | √ |  |  |  |
| | Huang, J. et al. 2017 | √ |  |  |  |  |  |  |  | √ | √ |  | √ |  |  |
| | Jha et al. 2019 | √ |  |  |  |  | √ |  |  | √ |  |  |  | √ |  |
| | Kumar et al. 2019 |  |  |  |  |  |  |  |  | √ |  |  | √ |  |  |
| | Li et al. 2016 | √ | √ | √ |  |  |  |  | √ |  |  | √ |  |  | √ |
| | Mai et al. 2018 | √ |  | √ |  |  |  |  | √ |  | √ |  |  |  |  |
| | Saifee et al. 2019 | √ |  |  |  |  |  |  |  | √ |  |  |  |  |  |
| | Samtani et al. 2017 |  | √ |  |  |  |  |  |  |  | √ | √ |  |  |  |
| | Siering et al. 2016 |  | √ |  |  |  |  |  |  |  | √ | √ | √ |  |  |
| | Shi, D. et al. 2017 |  |  |  |  | √ |  | √ | √ |  | √ |  | √ |  |  |
| | Van Osch et al. 2018 |  |  | √ |  | √ |  |  |  |  | √ |  |  |  |  |
| | Velichety et al. 2019 | √ |  | √ | √ |  |  |  | √ |  | √ |  |  |  |  |
| | Wang, Z. et al. 2020 | √ |  | √ |  |  |  |  | √ |  | √ | √ | √ |  |  |
| | Yoo et al. 2019 |  |  |  |  |  |  | √ |  |  | √ |  |  |  |  |
| | Zhang, D. et al. 2016 |  | √ |  |  | √ |  | √ | √ | √ | √ |  |  |  |  |
| | Zhou et al. 2018 |  | √ |  |  |  |  | √ | √ | √ |  |  |  |  |  |
| **MIS Quarterly** | Abbasi et al. 2018 | √ | √ | √ |  | √ |  | √ |  |  |  |  |  |  |  |
| | Arazy et al. 2020 |  | √ |  |  |  |  | √ |  |  |  |  |  |  |  |
| | Bapna et al. 2019 |  | √ |  |  |  |  |  |  |  | √ |  |  |  |  |
| | Benjamin et al. 2019 |  |  | √ |  |  |  |  |  |  | √ | √ |  |  | √ |
| | Chau et al. 2020 | √ |  | √ |  | √ | √ |  | √ |  | √ |  | √ |  |  |
| | Deng et al. 2018 | √ |  |  |  | √ |  |  |  |  | √ |  | √ |  |  |
| | Gong et al. 2017 |  | √ | √ |  |  |  |  |  |  | √ |  | √ |  |  |
| | Huang, K.Y. et al. 2019 |  |  |  |  | √ | √ |  |  |  | √ |  | √ |  |  |
| | Huang, N. et al. 2016 | √ | √ |  |  | √ | √ |  | √ | √ |  |  |  |  |  |
| | Li, J. et al. 2020 | √ | √ | √ |  | √ |  |  |  |  |  | √ | √ |  |  |
| | Liu, X. et al. 2019 |  | √ |  |  | √ |  |  |  |  | √ | √ |  | √ |  |
| | Rhue & Sundararajan 2019 | √ |  |  |  |  |  |  |  | √ | √ | √ |  |  |  |
| | Shi, Z. et al. 2016 |  | √ |  |  |  |  | √ |  |  | √ |  |  |  |  |
| | Shin, D. et al. 2019 | √ | √ | √ |  |  | √ |  | √ |  | √ | √ | √ |  |  |
| | Wu, J. et al. 2019 | √ |  |  |  |  |  | √ | √ | √ |  | √ | √ |  |  |
| | Yue et al. 2019 | √ | √ |  |  | √ |  | √ | √ |  | √ |  |  |  | √ |
| | Zhang, K. et al 2016 | √ |  |  |  |  | √ |  | √ |  | √ |  |  |  |  |
| | Zhang, W. & Ram 2020. |  | √ |  |  | √ | √ |  |  |  | √ |  |  |  |  |

| Key | √ = less than full utilization | √ = full utilization | Using word representation |
|---|---|---|---|

Table II: Text Mining Techniques and Their Applications in IS Research

| Text Data Source | Methods | | | | |
|---|---|---|---|---|---|
| | Sentiment Analysis | Emotion Detection | Text Classification | Topic Modeling | Word Vector Models |
| Reviews | ✓ | ✓ | ✓ | | ✓ |
| Social Media | ✓ | ✓ | ✓ | ✓ | ✓ |
| Other Documents | | | | ✓ | ✓ |
| Apps | | | | ✓ | ✓ |
| Other Web Content | | ✓ | ✓ | ✓ | ✓ |
| Key | ✓ = used in IS literature | | ✓ = used feature extraction | | |

## 4  IMPLICATIONS FOR RESEARCH AND PRACTICE

TLMs will enable researchers and practitioners to leverage the broad NLP power of transformers for task-specific applications through DTL via fine-tuning and through the development of custom models with rich CWRs. In doing so, they offer opportunities to improve insights from existing research and they will open the door to powerful NLP-driven analyses across a wide variety of new domains and applications. Furthermore, they will enable fundamental changes in the nature of human-computer interaction by enabling the widespread use of LUIs. While it is not possible to anticipate the upper bound of the technical capabilities that TLMs will unlock, much less the ways in which TLMs will impact organizations and society, we still attempt to outline those ways that seem plausible based on the bodies of literature we have reviewed in this survey.

In the previous sections we have only considered the technical elements of TLMs and the existing body of IS literature that utilized text mining. In this section we further explore recent literature regarding TLMs, but with a focus on their applications in both research and practice. We do this in two phases: 1) by examining their potential for use in traditional text mining/NLP tasks and applications, and 2) by examining their potential for use on NLP tasks for that have not previously been practically useful due to insufficient performance. Through each of these phases we consider tasks and applications through their standard typification in the text mining and NLP literature. Throughout this process we consider implications of our discussion on both research and practice, especially in areas where they can be used to enhance or broaden the body of existing IS research[15].

### 4.1  Enhancing Existing Text Mining and NLP Applications

#### 4.1.1  Sentiment Analysis

Sentiment analysis is the most widely used text analytics technique in recent IS literature (see Table I) with a broad range of applications including ecommerce, market intelligence, social media analytics, government, politics, security and public safety [Chen 2012]. TLMs have already been used to improve upon the state-of-the-art (SOTA) performance on benchmarks for the major classes of sentiment analysis: aspect-based sentiment analysis, fine-grained sentiment analysis, targeted sentiment analysis and emotion detection [Phan 2020; Cheang 2020; Naseem 2020; Zhong 2019]. However, while we do feel strongly about the potential for TLMs, we also recognize sentiment analysis is a complex and well-established field and we do not intend to suggest that TLMs can nontrivially be used to improve on all existing applications. A comprehensive discussion of the implications of TLMs on sentiment analysis in IS research is beyond the scope of our study, but in this subsection we highlight the ways in which we feel that these techniques can be applied to benefit researchers and practitioners. Due to the prevalence of sentiment analysis in IS research, we do not provide many explicit examples but rather focus on the potential of the latest developments.

---

[15] We note the caveat that, while many of the studies cited here produce state-of-the-art results, their results may not yet enable new forms of research as we suggest. However, we feel our suggestions are prescient and justified due to the rapid pace of progress, and due to the fact that most of the studies cited utilize BERT [Devlin 2019], which only represents the baseline in the SuperGLUE [Wang 2019a] benchmark.

For general applications, XLNet [Yang 2019] has demonstrated SOTA performance for the largest variety of benchmarks and is the best suited TLM for fine-tuning tasks involving task-specific sentiment analysis. More recently, the TLM SentiLARE has demonstrated strong all-around performance in sentiment analysis tasks by incorporating linguistic knowledge from SentiWordNet (Ke 2020). TLMs and CWRs have also been used to outperform LSTM and SVM-based methods on investor sentiment analysis [Li 2020] and to achieve SOTA results on targeted, domain-specific datasets such as airline industry Twitter data [Naseem 2020]. Implementations of TLMs such as these have strong implications for IS researchers utilizing sentiment analysis on domain-specific, targeted topics.

Some have gone as far as to suggest that BERT [Devlin 2019] be used as the standard baseline for comparing future progress [Li 2019]. We believe that this is reasonable, and suggest that widely used TLMs (*e.g.*, BERT) should be used as baselines for comparing all relevant novel text mining or NLP methods moving forward. We further suggest that due to concerns about the impact of mismeasurement and misclassification error from extracted data mining features on the validity of IS research [Yang 2018] those who choose to use alternative sentiment analysis techniques as input features for statistical models offer better justifications for their selected methods, including explanations for why fine-tuned TLM models were not used or including comparisons to more advanced TLM-based models[16].

Aspect-based sentiment analysis (ABSA) has great potential for business applications, such as for understanding online reviews in a finer-grained manner [Huang 2020], but our literature review indicates that it has not been applied in research published in premier IS journals yet. However, TLMs may make this easier in future work. DomBERT has been proposed to do just this (and more) by helping to train domain-specific TLMs with minimal resources, and it has shown promising results on ABSA tasks [Xu 2020]. Other recent work on improving analysis of online reviews has advanced the SOTA on widely used ABSA benchmarks of online reviews [Phan 2020], and we feel that using TLMs for ABSA is an excellent avenue for future IS research.

Due to the value of DTL, TLMs make targeted sentiment analysis a particularly easy area for improving upon the analyses of existing IS research when large amounts of data are available. Even in cases where labeled training datasets are not available, crowd sourcing options such as the Amazon Mechanical Turk make the labeling of a modest number of documents[17] reasonable. This is particularly useful for cases involving text data, like Tweets, that does not conform to standard syntactic rules, or cases when specific topics are of interest. For an IS example, Ghiassi et al. [2017] developed a custom sentiment model using feature engineering, vectorization and a trained classifier. However, TLMs remove the text mining knowledge necessary for developing an advanced model like this and make it more straightforward to achieve optimal performance on business related datasets, such as that used by Ghiassi et al., with only data collection, cleansing and sentiment scoring. For the same reasons, TLMs offer benefits to practitioners, where marginal improvements in the quality of input features and the quality of results can have a more tangible and valuable effect than in research by increasing revenue, sales or profits.

---

[16] Such as fine-tuned XLNet [Yang 2019] or SentiLARE [Ke 2020]. See Phan and Ogunbona [2020] for an example.
[17] A number of documents on the order of magnitude from 1,000 to 10,000 is often sufficient for fine-tuning pretrained TLMs.

### 4.1.2 Emotion Detection

Emotion detection is a form of sentiment analysis that we feel is worth mentioning separately due to its relevance to IS research. From text alone, it can be particularly challenging due to the absence of knowledge about the target's gestures or facial expressions [Chatterjee 2019a]. Progress on this task has proved to be more challenging than some of the other tasks discussed in this section where BERT-based models have easily improved upon SOTA results. However, progress has still been made, for example, by fine-tuning on evaluation datasets using TLMs that account for commonsense reasoning by incorporating a commonsense knowledge base and an emoticon lexicon during pretraining[18] [Zhong 2019]. Commonsense knowledge has also been employed more recently by using pretrained CWRs to incorporate different commonsense elements such as mental states and causal relations to learn interactions between interlocutors in dialogue, achieving SOTA performance on four conversational emotion benchmarks [Ghosal 2020]. Emotion detection is more challenging than other tasks that we have focused on, but these early results offer strong evidence for the ability of combining CWRs and TLMs with other techniques (*e.g.*, knowledge graphs) to outperform existing models on such challenging tasks.

While more complex, these solutions make progress on a topic that is particularly valuable in IS research and business more broadly. A number of studies published in elite IS journals involve emotion, however, they often involve designed experiments [Liang 2019] or qualitative and mixed-methods approaches [Salo 2020]. Consequently, we feel that methodological research involving TLMs for emotion detection is a topic that is well-suited for IS researchers and should be prioritized due to its potential impact in IS research and beyond. We further expect that emotion detection could have an impact on other areas of business research such as marketing[19] and finance[20].

In our IS literature review, Chau et al. [2020] demonstrated a novel model utilizing a text mining driven classifier in tandem with a rule-based classifier to identify at-risk individuals exhibiting emotional distress. This is a novel and important application of text analytics in IS research, yet, in light of the literature reviewed here there is much room for improvement[21], and it would be interesting to see the methods discussed in this section used for future work along these lines. The data used by Chau et al. was in Chinese, so some of the techniques suggested above may not have been viable alternatives, but multilingual TLMs, discussed later in this subsection, offer new solutions for this as well.

---

[18] This example differs from the majority in that it involved the specialized pretraining of a TLM as well as task-specific fine-tuning (as opposed to using an out-of-the-box TLM pretrained on a large, generic corpus).

[19] As an example, from marketing, Rockledge and Fazio [2020] examine the effects of emotion in online reviews using a lexicon-based emotion analysis technique. We feel that this could likely benefit from more fine-grained analysis using either ABSA-based methods or some of the more complex commonsense-based emotion detection techniques discussed above.

[20] In finance, studies focus on sentiment and emotion, but they do not use text mining techniques [Jiang 2019; Cortes 2016]. We feel that these new methods may be strong enough to lead to valuable insights which can aid in the development of new theoretical contributions, and we suggest that researchers and practitioners in these disciplines consider applying the methods discussed in this subsection for datasets available in their domain.

[21] For one, it uses a lexicon-based method for feature extraction, which is not relevant enough for comparison on emotion recognition benchmarks in our literature review or in the foremost computing psychology journal [Chatterjee 2019b]. We feel that it would have been useful for a study so recent to have included a comparison to the current methods discussed here.

### 4.1.3 Text Classification

Text classification is an essential technique of text mining that has numerous applications in organizations. While it was not as widely used in our survey as sentiment analysis or feature extraction, it was commonly used in combination with other text mining techniques and is one of the techniques which stands to improve most dramatically from fine-tuning TLMs. This is underscored by the fact that, in their seminal paper on transfer learning for language models, Howard and Ruder [2018] focused on six text classification tasks for demonstrating the value of DTL in NLP. BERT [Devlin 2019], fine-tuned on domain-specific datasets, was quickly demonstrated to achieve SOTA performance for a variety of text and document classification tasks [Yao 2019; Sun 2019b]. One example that could be particularly useful for IS research is BERTweet, which is a BERT-based model pretrained on Twitter data that achieves SOTA performance on Twitter text classification as well as part-of-speech-tagging and named-entity recognition [Nguyen 2020]. Models like this, pretrained on domain-specific data, are quite common: SciBERT [Beltagy 2019] and COVID-Twitter-BERT [Müller 2020]. Such models can then be fine-tuned on task-specific data for further performance gains. Due to the improved performance they bring, it is likely that similar models could be very useful for numerous applications in IS research, other business domains and the social sciences more broadly. For an IS example, one could extend the work of Mejia et al. [2019] on classifying restaurant hygiene by unsupervised pretraining of a BERT-based model on bulk restaurant reviews, then fine-tuning for classification of "instances of hygiene violations."

TLMs enable the creation of text classification models which previously required complex methods to be created with significantly less effort and expertise. Huang et al.'s [2020] study of support and companionship in virtual healthcare communities offers an excellent opportunity to use fine-tuning to improve model performance. BERT [Devlin 2019] could be fine-tuned via a Google Colab notebook[22] (and a powerful coprocessor[23]) for free[24], as could smaller T5 models [Raffel 2020]. However, Colab offers a good opportunity to debug T5, and Huang et al.'s study offers a good opportunity to use T5's multitask capability. As another example, Kraus and Feuerriegel [2017] developed a Bi-LSTM model for predicting a firm's market performance based on financial disclosures, but BERT could be fine-tuned on the same data using Colab and a few dozen lines of code to improve performance (see Wolf 2019).

Training a TLM to classify the data from Kraus and Feuerriegel [2017] would work by simply inputting entire documents because the model would simply output a class, but not all documents are short enough to fit in the context window of TLMs[25]. Innovative models such as DocBERT [Adhikari 2019] and the Longformer [Beltagy 2020] have achieved SOTA results on various document classification tasks, as well as other document related tasks, and could be useful for longer documents like internal reports, legal documents, newspaper and magazine articles or longer Wikipedia articles. Moreover, recent modifications to the original transformer architecture such as the reformer [Kitaev 2020] suggest that larger context windows will be a feature of TLMs in the near future.

---

[22] Google Colab notebooks (virtual Jupyter notebooks) can be used to run simple deep learning models directly through the browser, and we feel that these are well suited for most applications of TLMs and CWRs suggested in this study. Official instructional notebooks exist for all of the most widely used models, and 3rd party notebooks exist for many other models.
[23] This will either be a SOTA graphics processing unit (GPU) or one of Google's proprietary tensor processing units (TPUs).
[24] Colab is free but has usage limits. However, Colab Pro, for $10 per month has no limits and more generous TPU allocation.
[25] There is a limit to the number of tokens that can be input (*e.g.*, 512 for BERT) [Devlin 2019], though this context window is larger for larger models such as T5 (*e.g.*, up to 2,048) [Raffel 2020].

### 4.1.4 Topic Modeling

Topic modelling is widely used in IS research, as indicated by our survey, and the most widely used technique is latent Dirichlet allocation (LDA) [Blei 2003]. However, TLMs have also performed well in these areas and BERT [Devlin 2019] has been shown to improve upon the SOTA when applied to specific use cases such as argument [Reimers 2019] and document clustering [Park 2019]. Moreover, contextual document embeddings from TLMs have been shown to improve topic coherence [Bianchi 2020]. However, overall it is unclear whether BERT-based CWR clustering improves on LDA enough to make a difference [Sia 2020], but the results from Sia et al. suggest that larger TLM CWRs such as those from RoBERTa [Liu 2019c], XLNet [Yang 2019] or T5 [Raffel 2020] could be expected to outperform LDA. While there may be some uncertainty about using CWRs for clustering, Hoyle et al. [2020] have demonstrated that TLM-based techniques can be used to obtain SOTA topic coherence. They do this not by using CWRs or TLMs directly for topic modeling, but by using their BERT-based Autoencoder Teacher (BAT) approach in tandem with SOTA topic modeling methods. Thus, this is another case in which TLM-based methods should begin to be used as default methods. This can have important implications for IS research because improved input features can significantly impact the statistical validity of IS research results [Yang 2018].

### 4.1.5 Word Representation Models

Word representation models are commonly used in IS research, particularly when feature extraction is necessary. While such techniques do not outperform CWRs, they are still able to perform relatively well on tasks with plentiful data and simple language [Arora 2020], but our review has indicated that this is not always the case for IS. Thus, we see numerous studies as being able to benefit from the improvements offered by CWRs. For example, Arazy et al. [2020] focus on the evolution of digital artifacts (*i.e.*, wiki articles) over time by tracking trajectories in a feature space, and, because the authors do not use text mining, they explicitly suggest the use of word representations would benefit future work[26]. As another example we consider Wang et al. [2020] who extract soft semantic factor characteristics from descriptive loan texts, but the semantic similarities between words and loan texts could be more easily and effectively captured in a latent feature space using CWRs. Numerous other studies utilize feature extraction, some even using neural word representations, and many stand to gain from using more advanced CWRs (see Appendix A for concrete suggestions on the papers from the IS literature review).

Other models in the IS literature have used alternative techniques for feature extraction to develop novel distributional representations of text [Shi 2016], and we feel that some of these models offer good opportunities for using CWRs. For example, Shi et al. used LDA [Blei 2003] for feature extraction to represent aspects of firms' business to evaluate firms' relative "business proximity." Lee et al. [2020] also use LDA to create a novel "app similarity measure." Work such as this is well suited for CWRs which can be crafted in a custom fashion to create novel measures of documents' semantic similarity [Gyawali 2020]. In general, LDA is widely used in the IS research literature for extracting features [Gong 2018; Shin 2019; Liu 2020b], but even if dimensionality reduction is necessary for using the features in statistical models, we agree with Shin et al. that CWRs can provide richer representations.

---

[26] Specifically, they suggest that "the feature space could be represented through more sophisticated text processing methods and more advanced knowledge representations … the patterns observed here serve as a lower bound" [Arazy 2020].

## 4.2 Beyond Existing Applications

While CWRs and TLMs have significant implications for improving and furthering existing IS research, we believe that their most interesting applications for IS research are in their advanced and novel applications. In this subsection we discuss these emerging topics.

### 4.2.1 Regression

Regression is one emerging application for which little previous work has been conducted in NLP. One good example of using NLP for regression was by Kraus and Feuerriegel [2017] who used financial disclosures to predict firms' subsequent performance in financial markets, but their model required a very specialized LSTM model. However, it is possible to simply fine-tune language models for regression by posing regression problems as text-to-text tasks [Raffel 2020]. While, this is still an emerging research area, it has been demonstrated for applications such as table retrieval [Chen 2020] and to predict brain activity as measured by fMRI based on the text being read [Schwartz 2019]. One practical example of regression on text data is that of automated essay scoring such as for standardized tests. Yang et al. [2020] find that simply fine-tuning on BERT [Devlin 2019] is not enough, but that extracting CWRs from BERT and training a fully-connected neural network on multiple losses improves on the SOTA performance by almost 3%. We feel that this example offers promise for many practical business tasks, as well as for numerous uses in IS research.

### 4.2.2 Multilingual Analytics

Machine translation can be useful in business intelligence and business analytics applications when organizations need to analyze or monitor either static or streaming text data in multiple languages [Moreno 2016]. While it is more commonly thought to be an independent research area within NLP, like speech recognition, progress on the related topic of multilingual language models does have significant implications for IS research. Machine translation has more applications in practice than research, and interested readers are encouraged to review recent high-level overviews [Hao 2019]. Here, our discussion focuses broadly on multilingual capabilities of TLMs and their applications in both research and practice.

Multilingual TLMs were introduced in the previous section and models like XLM-R [Conneau 2020] result in significant improvements for a wide variety of cross-lingual transfer tasks. What is most interesting about the results from Conneau et al. is that they suggest these gains may be possible without sacrificing monolingual performance. It may not be obvious how this will impact IS research, but there is a digital language divide between dominant languages [Young 2015], and as information technology has proliferated over time this divide has had a significant impact on their adoption and applications across cultures. Consequently, multilingual TLMs enable a powerful tool to examine this using recent deep learning analytics. For an IS example, George et al. [2018] evaluated the effect of communication media and culture on deception detection by conducting an experiment which showed that different combinations of media and cultural effects affected deception detection accuracy. Multilingual language models offer the ability to conduct research in this vein without the burden of conducting an experiment with groups across three different languages, a burden that is likely prohibitive to most IS researchers. Our survey of IS literature found social media and online reviews to be the primary applications of text mining in IS research. Simply considering social media, and the ability to apply TLMs for analysis of behavior across cultures, one can look at recent work in leading human-computer interaction journals [Wang 2019b; Cho 2018] and foresee the strong research potential here. Thus, we anticipate that multilingual TLMs will open doors to numerous new research directions for IS researchers.

Yet, these models' value is not limited strictly to cultural comparisons and can be applied directly to improve insights from existing IS research. The work of Chau et al. [2020] mentioned earlier could benefit from using multilingual representations to replace older lexicon-based methods of feature extraction. This hints at the possibility of being able to conduct IS research on non-English datasets without the need for fluency in the language of focus. If possible, this would open up a wide variety of foreign language datasets to IS researchers.

### 4.2.3  Language Generation

Language generation has been a topic of interest in the NLP community for over a decade, and it is such a significant topic with respect to TLMs that we make a distinction between standard TLMs and generative language models. GPT-2 [Radford 2019] was the first generative language model to really demonstrate shockingly impressive language generation results. It was followed by T5 [Raffel 2020] and most recently by GPT-3 [Brown 2020], which each demonstrated shocking gains.

Significant effort is going into improving reliability and ease of generating samples that are more human-like [Keskar 2019] or less biased [Huang 2020; Ma 2020] while others are focusing on applying TLMs to more immediately practical applications, such as chatbots [Roller 2020b]. We previously mentioned chatbots that were closing in on human-level performance for open domain conversation [Adiwardana 2020; Roller 2020a]. We expect language generation to be inextricably related to the future of IS research in a very significant way given its potential to fundamentally change human-computer interaction. The remainder of this section focuses on different applications of language generating systems with implications for future IS research such as for document summarization, question and answering, automated report generation and language user interfaces.

### 4.2.4  Document Summarization

Document summarization is a task that has the potential to be very valuable for business intelligence and business analytics applications. While document summarization is still a very challenging task [Kryściński 2019], TLMs are showing promise in this area, and have even successfully been able to use recursive summarization schemes for summarizing entire novels [Wu 2021]. Generally, document summarization is classified as being one of two types: extractive or abstractive. Extractive summarization involves identifying and concatenating extracts from the document into a summary. Improvements for this using TLMs are straightforward for specialized applications because existing pre-trained models can simply be fine-tuned on domain-specific datasets [Gu 2019]. However, abstractive summarization is more challenging, yet, despite this, more complex TLMs have been able to achieve SOTA performance when trained directly on task-specific datasets [Duan 2019]. Researchers have begun using unified frameworks for multitask models capable of both abstractive and extractive summarization [Chen 2019], leading to SOTA on benchmarks for both extractive and abstractive summarizations [Liu 2019b] and multi-document summarization [Jin 2020]. Abstractive summarization is more valuable in the long run, and recent work on this task has concluded that TLMs and generative language models are able to generate more informative, coherent, faithful and factual summaries [Maynez 2020].

The potential applications of summarization for business intelligence systems are wide-ranging. For one, if we extend summarization to full report generating systems, we can envision how such systems could leverage industry reports, news articles and social media to power business intelligence systems with real-time understanding of complex market behavior in the form of an intelligent dashboard. Summarizations could also be used for reducing reading time on emails or other long documents or reports produced by employees at all levels of the organization. The ability to highlight the key points in a document may even be more beneficial in this aspect. Exciting new work from OpenAI has shown significant improvements in summary quality by using human feedback to train summarization models [Stiennon 2020] and these results suggest that practical use of summarization systems may not be far away.

Multi-document summarization [Lu 2020] and extreme summarization [Narayan 2020] have become popular topics as well, and ones with significant implications for practical applications in highly specialized domains (*e.g.*, science, finance, etc.). Extreme summarization refers to summarizing highly technical documents, such as scientific papers, with a single sentence. This could also be very useful for summarizing financial statements or legal documents. Documents of this sort are often large in number, and query focused multi-document summarization that is effective for a range from coarse-to-fine estimation [Xu 2020] could be extremely useful in future business intelligence systems for these domains.

Another potentially very useful application of summarization would be practical cross-lingual summarization, which would use a multilingual TLM to generate a summary in one language from a text written in another language. TLMs have been used for this, but the relative performance was not possible to determine [Zhu 2019]. Exciting new work shows continued progress on this task [Cao 2020], and a new multilingual summary dataset [Scialom 2020] and benchmark [Ladhak 2020] suggest that we can expect more work on this topic in the future. Similar to our discussion of multilingual TLMs earlier, the applications discussed in this subsection offer numerous opportunities for IS researchers and open the door to novel research questions.

### 4.2.5 Question Answering

Question answering (QA) systems that are effective for domain-specific applications have tremendous potential for business intelligence. Such systems could fundamentally change the nature of decision support for any application with enough data for fine-tuning. Impressively, systems have been able to score an A on a standardized New York 8[th] grade science exam and a B – an *83* – on the same 12[th] grade exam [Clark 2019]. While this sort of generality is not necessary for practical applications, it effectively demonstrates how powerful QA systems from TLMs can be. For many practical business applications, a high school graduate that can make an *83* on the most difficult standardized high school level science exam can likely read documents and be able to generate answers that suffice for a wide range of data and applications relevant to organizations. Many tasks that standard college graduates do in white collar jobs do not require the full use of their faculties and education.

Such powerful QA systems, particularly when it is possible to fine-tune them for customized experiments, have the potential for valuable new directions in IS research. For example, we can consider QA systems that are easily fine-tuned on domain-specific datasets. This has been a desirable goal for many years, especially since IBM's Watson, but it has not materialized as many had originally anticipated. Yet, given the rapid progress of TLMs, we can expect such systems to become practical in the near future. Xu and Lapata [2020] discuss adapting recent QA methods to improve query focused multi-document summarization, and such systems have the potential to transform strategic decision making in organizations and dramatically impact the nature of white-collar labor. If we consider not just multi-document QA, or data warehouse QA, but QA based on an entire organization's archived text data, we can begin to understand this potential. Yet, however transformative these technologies may be, it is likely that these advances will first lead to the augmentation of human jobs rather than the replacement of them [Morgan 2019], and it falls on IS researchers to develop an understanding of how this augmentation of occupations will impact organizations and the future of white-collar work.

Recent work on a knowledge-intensive generative language model from Facebook – retrieval-augmented generation (RAG) – demonstrated SOTA performance for three widely used, general QA tasks [Lewis 2020c]. In the same month, another QA oriented generative language model from the Allen Institute, called UnifiedQA, demonstrated strong performance without fine-tuning, and was able to achieve SOTA performance on 10 factoid and commonsense QA benchmarks [Khashabi 2020]. However, while all of this may seem impractical due to the lack of labeled datasets, there are strong, user-friendly extractive QA systems that can be fine-tuned on large, unlabeled domain-specific datasets [Dibia 2020] which could be used for IS research now and which could offer guidance for future research as business-related question answer datasets are created and as systems grow more capable[27].

Generally, work on reading comprehension is closely related to QA systems. Thus, it should come as no surprise that CWRs were already rivaling SOTA performance in related tasks in 2018 [Salant 2018]. After its release, BERT [Devlin 2019] soon achieved new SOTA performance on multiple benchmarks in multiple choice reading comprehension tasks [Zhang 2019]. Based on the prevalence of online reviews that our survey illuminated in existing IS research, the new idea of *review reading comprehension* proposed by Xu et al. [2019], for including a QA system on top of a large repository of ecommerce reviews, may offer some further insight into the potential for fine-tuning multitask models on domain-specific datasets. Their system targeted customers, but similar systems could be developed for other applications in organizations such as for analysts working to increase revenue or to improve customer satisfaction.

### 4.2.6 Language User Interfaces

Language user interfaces (LUIs) have long been anticipated to become a widely used modality of human-computer interaction [Brennan 1991]. While LUIs still play only a limited role in our daily interactions with computers, recent progress in TLMs raises the possibility that LUIs will become practical and widespread in the near future. We envision practical LUIs to be powerful systems that are used to enhance human capabilities through human-computer interaction [de Vries 2020]. In the following paragraph we will briefly discuss some possibilities for practical applications of LUIs[28].

*We define an LUI as an intelligent system that is goal oriented to substantially enhance economically valued human capabilities through an interface that is optimally controlled with natural language.* We are particularly interested in LUIs that are practical in the sense that they can assist humans in tasks of nontrivial economic utility. Some obvious examples of LUIs are personal assistants, assistants for the impaired or customer support assistants. While many call centers already use automation and there are widely used personal assistants like Google Assistant and Siri, their economic utility is relatively limited. Perhaps this is truer for the personal assistants than for the call centers but call center automation has been gradually increasing for decades. Many tasks are repetitive, like navigating information systems, and they do not require strong language understanding or interaction. Thus, such systems do not meet the criteria of being optimally controlled through natural language. Furthermore, while we do not feel any of these existing example systems meet the criterion of

---

enhancing economically valued human capabilities, we feel that TLMs are currently poised to usher in dramatic progress on it.

More powerful LUIs that we foresee include navigation agents for automobiles or flying vehicles, interactive domestic appliances and domestic robots. Furthermore, directly related to organizations' productivity, we envision agentive business intelligence systems that are able to offer powerful capabilities such as those mentioned earlier in this subsection like summarization or QA capabilities, but which also leverage reinforcement learning to tailor their functionality to a specific user. Such systems would truly transform the nature of business intelligence and decision support systems, and it is critical for IS researchers to begin understanding how these systems will change organizations and society in the years to come because their rise may come quickly [Gruetzemacher 2020].

While this is primarily a topic for future research, it is possible for eager IS researchers to begin work on these problems at present. We have included links (in footnotes) in this subsection to open-source code that could be used to these ends. Further, Roller et al. [2020a][29] demonstrated the best performing chatbot[29] to date in their Blender Bot while also releasing the code open-source as well as the 9.4 billion parameter pretrained model[30]. We believe that this alone unlocks a wide range of novel IS research directions, particularly if the model is fine-tuned for specific tasks and evaluated empirically. Other recent work on ToD-BERT [Wu 2020] for task-oriented dialogue offers another tool[31] for conducting preparadigmatic research in this new domain. We feel that such research is important because children are already accustomed to LUIs like Alexa and Siri in their homes and phones and are beginning to expect devices to respond to verbal commands; we anticipate that in the coming decades, when entering the workforce they will expect language-enabled support in the workplace.

### 4.2.7 Few-Shot Learning

Few-shot learning is something that we feel will be closely related to LUIs, but its potential impact has strong enough potential to garner a brief but independent discussion. The strong performance of GPT-3 on certain tasks such as QA via zero- one- and few-shot learning suggests the possibility of novel LUIs, and we feel that this is a topic that also falls to IS researchers to explore. It is beyond the scope of this study to illuminate in detail the potential for few-shot learning in IS research, but we suggest considering the following. Generative language models like GPT-3 take a text prompt at the time of inference, and, in the case of GPT-3, this prompt can be long and involve sequential tasks such as questions followed by answers. Performance from GPT-3 on tasks demonstrated in this manner is particularly strong, as we discussed in an earlier section.

GPT-3 also performs well on prompts that give a context and ask for the model to fill in the blank, to complete the sentence or even to generate an essay based on the prompt. Thus, it is easy to see how continued research and increasing the scale of powerful generative language models like GPT-3 can lead to very useful systems capable of report generation, summarization and other valuable tasks if trained with a larger context window (and at a higher computational cost). However, what is not as obvious is the value of direct user interface with a system capable of learning complex contexts such as QA or mathematical operations. It is likely that there are novel ways of interacting with such interfaces that can create value in ways that are difficult to imagine

---

[29] While there has been dramatic and practical progress on open domain chatbots recently, a full discussion of chatbots is beyond the scope of this study. (This topic will be covered in greater detail in forthcoming work from the authors on LUIs.)
[30] This can be found at: https://parl.ai/projects/recipes/.
[31] The code can be found at: https://github.com/jasonwu0731/ToD-BERT.

*a priori.* For example, one startup is using GPT-3 exclusively to improve inbox productivity by generating detailed emails from short prompts[32]. They do this through a novel notion of LUI wherein the user does not have to reply in complete sentences, they only provide the information necessary for the response and the model generates an email response in context[33] with the correct information. We feel this is an appropriate and urgent topic for IS research[34].

## 5  SUMMARY OF IMPLICATIONS FOR IS RESEARCH

We first surveyed the recent progress in NLP that has led to SOTA performance on a wide range of tasks using TLMs. Next, we discussed and reviewed IS research that has used existing text mining and NLP techniques, a substantial portion of which could be improved by using CWRs or TLMs. We then discussed some of these possible improvements in the next section[35] as well as a number of possible avenues for new and novel IS research stemming from TLMs. In this section we summarize our findings and their implications for IS research.

TLMs have a handful of distinct and noteworthy advantages over standard text mining techniques. Foremost, they are able to achieve SOTA results on a wide variety of text mining and NLP tasks as long as a modestly sized dataset is available for training. However, their value is not limited to labeled datasets and fine-tuning; they can be used to generate rich CWRs which can be used to extract features for building custom models in combination with a variety of machine learning or statistical methods. Based on how often feature extraction has been used for text analytics in the recent IS literature, we feel that this alone can have a significant impact on future IS research.

The remainder of this section focuses of four distinct topics. First, we review the implications of TLMs which is followed by a discussion of new opportunities and LUIs. We next offer some brief suggestions for reviewers and editors when considering submissions involving novel methodological contributions for text analytics and finally, we discuss some implications of further TLM scaling and continued progress in grounded semantics.

### 5.1  Transformer Language Models (TLMs)

From sentiment analysis to emotion detection to text classification to regression to cross-lingual analysis, TLMs promise to have a significant positive impact on future IS research, notwithstanding the more advanced novel applications, and they can do so in a number of different ways. For improving existing work, they can either 1) be used to generate rich CWRs or 2) be used directly, through pretraining, through fine-tuning and DTL, or both. More exciting, they can 3) be used to extend existing IS research by enabling easier cross-cultural analyses. We describe these themes in this subsection and discuss other issues that may impact the future use of TLMs.

1. When a modestly sized dataset is available, simple DTL and fine-tuning will often outperform all methods other than specialized TLMs or advanced models using CWRs and possibly LSTMs. This is important because using DTL to obtain such strong performance is significantly easier than the development of a custom LSTM model or the development of a custom TLM model or one that has not been pretrained. This should enable the wider use of fine-tuned TLM models for tasks such as sentiment analysis or text classification, thereby improving performance, even if only as a component of a more complex analysis.

---

Because Google offers free, powerful Colab notebooks that include tutorials for fine-tuning a standard BERT model[36], we feel it is reasonable to expect IS researchers to be able to do this with minimal machine learning expertise[37].

2. CWRs generated from TLMs are superior for feature extraction from large datasets which are able to support high-dimensionality machine learning models. When this is not possible, CWRs can still be very valuable for feature extraction when coupled with dimensionality reduction and feature selection techniques. Due to the prevalence of feature extraction in our survey of IS literature, we feel that CWRs should be more widely used as it stands only to benefit IS research by reducing bias from mismeasurement [Yang 2018].

3. Multilingual TLMs enable DTL to leverage multilingual representations for cross-lingual analytics. While this is still an emerging topic in NLP research, models such as XLM-R [Conneau 2020] are able to maintain monolingual performance while also being able to generate valuable representations for other languages. This can be very useful for IS research because it enables analysis of the effects of culture on social media behavior, technology acceptance, technology use, etc., all without having to design an experiment involving multiple languages. Moreover, it enables the use of SOTA methods when working on datasets involving foreign language data so that research in elite IS journals does not have to rely on older, more rudimentary methods [Chau 2018].

### 5.2 Novel Applications & LUIs

LUIs have been introduced as an incredibly promising area for IS research due to the recent progress in TLMs. While a full discussion of TLMs is beyond the scope of this paper, it is easy to see a path toward LUI research already emerging in the form of strong pretrained chatbots such as BlenderBot [Roller 2020a]. This chatbot is available open source[38], including the pretrained 9.4 billion parameter model, which enables IS researchers to begin working directly on LUIs.

Our literature review revealed few text analytics systems using design science, but LUIs will offer novel opportunities for using design science for artifact development and theorizing. Samtani et al. [2020] offer a strong template for such work, and we suggest that interested parties refer to it. We feel that this is a very strong potential area for research, and, because it is possible given existing technology, we suggest that interested IS researchers act fast to establish first mover advantage. We are eager and excited to see where research in this direction takes us.

### 5.3 Guidelines for Methodological Novelty Using TLMs and CWRs

TLMs offer a huge opportunity for researchers to improve upon previous SOTA results and to apply powerful NLP models to a wide variety of new applications which were previously not possible. With respect to improving upon SOTA, the ability of TLMs to do this is often related to the novelty and size of a training dataset rather than the novelty and methodological contributions of a technique. Thus, we encourage reviewers to be weary of papers employing TLMs which claim to make methodological contributions and to continue to seek theoretical contributions from novel studies using TLMs. However, this is not to say that methodological contributions cannot be made involving TLMs, but we suggest that it is necessary to compare results from proposed novel methods, like that of Chau et al. [2020], to static word representations as well as widely used CWRs. We further suggest that if TLMs or CWRs are used as a component of a proposed methodological contribution, the methodological contribution be made clear and robustly justified (*e.g.*, specialized pretraining such as for Zhong et al. [2019]). The deep learning IS research template of Samtani et al. [2020] is also useful for this.

---

[36] See: https://colab.research.google.com/github/tensorflow/tpu/blob/master/tools/colab/bert_finetuning_with_cloud_tpus.ipynb.
[37] Recall that pretraining schemes have a significant impact on downstream tasks: autoencoding models excel at discriminative tasks, autoregressive models excel at generative tasks, and sequence-to-sequence models attempt to balance performance between generative and discriminative tasks. Also recall that further fine-tuning and pre-finetuning can be utilized to enchance performance on many domain-specific tasks.
[38] See footnote 24.

## 5.4 TLM Scaling and Grounded Semantics

AI practitioners anticipate a continued trend in the scaling of computational resources to continue to drive progress in AI research for the next decade [Gruetzemacher 2020]. Taken with recent research from OpenAI [Kaplan 2020; Brown 2020] this suggests that TLM progress will continue to improve dramatically, but that the costs of this increased performance may be non-trivial and may make research and operationalization of the most powerful TLMs quite costly, possibly even cost prohibitive. As noted, OpenAI has already begun licensing an API for the largest GPT-3 model through Microsoft; the prices are anticipated to be extreme for fine-tuning but more reasonable for few-shot learning. However, it is also difficult to anticipate how quickly progress to increase language model efficiency, such as adapters [Houlsby 2019], might progress and interact with the AI practitioner forecasts for scaling.

If others follow the API licensing model, it has the potential to dramatically impact the future use of TLMs in both positive and negative ways. Most obviously, it could put TLM research and operationalization out of reach for many academics and firms, at least for lower priority projects. Alternately, the high cost of the service demands a user interface that ensures users will not waste their time with the API and incentivizes the provider to make the product easy to use and maximally effective. This could significantly impact firms' adoption of TLMs as well as their use in research and is an interesting research question for future work.

Continued progress in grounded semantics could also have a dramatic impact on the performance and practicality of TLMs. We feel strongly that higher levels of grounding, such as embodiment and social [Bisk 2020], are certainly not necessary for language grounding to begin to start seeing practical applications. Again, it is difficult to anticipate how quickly progress may come, but it is likely that existing work, once refined, can have an impact on the use of TLMs in text analytics and for applications such as question answering or summarization.

## 6 CONCLUSIONS

In this work we reviewed two bodies of literature: 1) literature related to recent progress in NLP and 2) recent literature involving the application of text mining and NLP published in the top IS journals. While some of the technologies we have discussed may mature over an extended period of time, it is important for IS researchers to keep up with the SOTA and to incorporate it into research without haste. This is true for all methodological progress, but it is particularly important for TLMs, CWRs, multilingual TLMs and LUIs as they have the potential to drive novel forms of IS research and substantially alter human labor and organizational processes for which text data is significant component. Even if such technologies are not mature, it is important for IS researchers to preemptively develop theory and methods for researchers and practitioners to use when the technologies do mature. We feel strongly that the IS research community, by more closely following progress in the NLP domain, can enhance the quality and value of their research contributions substantially. To these ends, we suggest the IS research community begin sponsoring workshops at the premier conferences (*e.g.*, NeurIPS, ICLR, ICML, ACL and EMNLP[39]) for business applications of these technologies[40].

Overall, the literature and the ensuing discussion led us to conclude that transformer language models are poised to dramatically reshape the use of text analytics and NLP in IS research and practice. Moreover, by enabling technologies such as language user interfaces, they are likely to precipitate transformative change in organizations and in society. Taken together, these topics offer significant opportunities for future work in IS research and we look forward to seeing what the future holds.

## ACKNOWLEDGMENTS

We thank Miles Brundage for comments on an earlier version of this manuscript. We also that anonymous reviewers from the 2020 Winter Conference on Business Analytics for pointing out the need for an independent survey paper on this topic.

## REFERENCES

Abbas, A., Zhou, Y., Deng, S., and Zhang, P., 2018. "Text analytics to support sense-making in social media: A language-action perspective," *MIS Quarterly, 42*(2), pp. 427-464.

Abbasi, A., Li, J., Adjeroh, D., Abate, M., and Zheng, W. 2019. "Don't Mention It? Analyzing User-Generated Content Signals for Early Adverse Event Warnings," *Information Systems Research, 30*(3), pp.1007-1028.

Adamopoulos, P., Ghose, A., and Todri, V. 2018. "The impact of user personality traits on word of mouth: Text-mining social media platforms," *Information Systems Research, 29*(3), pp.612-640.

Adhikari, A., Ram, A., Tang, R., and Lin, J. 2019. "Docbert: Bert for document classification," (*arXiv preprint arXiv:1904.08398*).

Adiwardana, D., Luong, M., So, D., Hall, J., Fiedel, N., Thoppilan, R., Yang, Z., Kulshreshtha, A., Nemade, G., Lu, Y., and Le, Q.V. 2020. "Towards a human-like open-domain chatbot," (*arXiv preprint arXiv:2001.09977*).

Aghajanyan, A., Gupta, A., Shrivastava, A., Chen, X., Zettlemoyer, L. and Gupta, S., 2021. Muppet: Massive Multi-task Representations with Pre-Finetuning. *arXiv preprint arXiv:2101.11038.*

Arazy, O., Lindberg, A., Rezaei, M., and Samorani, M. 2020. "The Evolutionary Trajectories of Peer-Produced Artifacts: Group Composition, the Trajectories' Exploration, and the Quality of Artifacts," *MIS Quarterly, 44.*

Arora, S., May, A., Zhang, J., and Ré, C. 2020. "Contextual Embeddings: When Are They Worth It?" *Proceedings of ACL 2020*, pp. 2650–2663.

Bapna, S., Benner, M.J., and Qiu, L. 2019. "Nurturing Online Communities: An Empirical Investigation," *MIS Quarterly, 43*(2), pp. 425-452.

Beltagy, I., Lo, K., and Cohan, A. 2019. "SciBERT: A pretrained language model for scientific text," *Proceedings of EMNLP 2019*, pp. 3615–3620.

Belz, A. 2019. "Deepfake News Generation: Methods, Detection and Wider Implications." Keynote address at the 12th International Conference on Natural Language Generation. Tokyo, Japan.

Bengio, Y., Ducharme, R., Vincent, P., and Jauvin, C. 2003. A neural probabilistic language model. *Journal of Machine Learning Research, 3*(Feb), pp.1137-1155.

Benjamin, V., Valacich, J.S., and Chen, H. 2019. "DICE-E: A Framework for Conducting Darknet Identification, Collection, Evaluation with Ethics," *MIS Quarterly, 43*(1).

Berant, J., Chou, A., Frostig, R., and Liang, P. 2013. "Semantic parsing on freebase from question-answer pairs," *Proceedings of EMNLP 2013*, pp. 1533-1544.

Bianchi, F., Terragni, S., and Hovy, D. 2020. "Pre-training is a hot topic: Contextualized document embeddings improve topic coherence," *Proceedings of First Workshop on Insights from Negative Results in NLP*, pp. 32-40.

Bisk, Y., Holtzman, A., Thomason, J., Andreas, J., Bengio, Y., Chai, J., Lapata, M., Lazaridou, A., May, J., Nisnevich, A. and Pinto, N. 2020. "Experience grounds language." *Proceedings of EMNLP 2020*, pp. 8718–8735.

Blei, D., Ng, A., and Jordan, M. 2003. "Latent Dirichlet Allocation," *Journal of Machine Learning Research*, 3(Jan), pp. 993-1022.

---

[39] The Conference and Workshop on Neural Information Processing Systems; The International Conference on Learning Representations; The International Conference on Machine Learning; The Annual Meeting of the Association for Computational Linguistics; The Conference on Empirical Methods in Natural Language Processing.

[40] Content at the first three conferences would not be restricted to NLP but could involve any applications of AI and machine learning in business. For this reason, one of these three conferences would perhaps be the best place to start.

Blohm, I., Riedl, C., Füller, J., and Leimeister, J.M. 2016. "Rate or trade? Identifying winning ideas in open idea sourcing," *Information Systems Research*, 27(1), pp.27-48.

Bommasani, R., Hudson, D.A., Adeli, E., Altman, R., Arora, S., von Arx, S., Bernstein, M.S., Bohg, J., Bosselut, A., Brunskill, E., Brynjolfsson, E., … and Liang, P. 2021. On the Opportunities and Risks of Foundation Models. *arXiv preprint arXiv:2108.07258*.

Brennan, S. E. 1991. "Conversation With and Through Computers," User Modeling and User-Adapted Interaction 1(1), pp. 67-86.

Brown, T., Mann, B., Ryder, N., Subbiah, M., Kaplan, J., Dhariwal, P., Neelakantan, A., Shyam, P., Sastry, G., Askell, A., Agarwal, S., Herbert-Voss, A., Krueger, G., Henighan, T., Child, R., Ramesh, A., Ziegler, D., Wu., J., Winter, C., Hesse, C., Chen, M., Sigler, E., Litwin, M., Gray, S., Chess, B., Clark, J., Berner, C., McCandlish., S, Radford, A., Sutskever, I., Amodei, D. 2020. "Language models are few-shot learners." *Advances in Neural Information Processing Systems.*

Brynjolfsson, E., and Benzell, S., and Rock., D. 2020. *Understanding and Addressing the Modern Productivity Paradox*, MIT Work of the Future Task Force.

Cao, Y., Liu, H., and Wan, X. 2020. "Jointly Learning to Align and Summarize for Neural Cross-Lingual Summarization." *Proceedings of ACL 2020*, pp. 6220–6231.

Chatterjee, A., Narahari, K.N., Joshi, M., and Agrawal, P. 2019. "Semeval-2019 Task 3: Emocontext Contextual Emotion Detection in Text," *Proceedings of 13th International Workshop on Semantic Evaluation*, pp. 39-48.

Chatterjee, A., Gupta, U., Chinnakotla, M.K., Srikanth, R., Galley, M., and Agrawal, P. 2019. "Understanding Emotions in Text Using Deep Learning and Big Data," *Computers in Human Behavior*, 93, pp. 309-317.

Chau, M., Li, T.M., Wong, P.W., Xu, J.J., Yip, P.S. and Chen, H. 2020. "Finding People with Emotional Distress in Online Social Media: A Design Combining Machine Learning and Rule-Based Classification," *MIS Quarterly*, 44(2), pp. 933-953.

Chaudhary, Y., Gupta, P., Saxena, K., Kulkarni, V., Runkler, T., and Shütze, H. 2020 "TopicBERT for Energy Efficient Document Classification," *Proceedings of EMNLP 2020*, pp. 1682-1690.

Cheang, B., Wei, B., Kogan, D., Qiu, H., and Ahmed, M. 2020. "Language Representation Models for Fine-Grained Sentiment Classification," (*arXiv preprint arXiv:2005.13619*).

Chen, H., Chiang, R., and Storey, V. 2012. "Business intelligence and analytics: From big data to big impact," *MIS Quarterly* 36(4), pp. 1165-1188.

Chen, K., Li, X., Luo, P., and Zhao, J.L. 2020. "News-Induced Dynamic Networks for Market Signaling: Understanding Impact of News on Firm Equity Value," *Information Systems Research*.

Chen, L., Baird, A., and Straub, D. 2019. "Fostering Participant Health Knowledge and Attitudes: An Econometric Study of a Chronic Disease-Focused Online Health Community," *Journal of Management Information Systems*, 36(1), pp.194-229.

Chen, W., Gu, B., Ye, Q., and Zhu, K. 2019. "Measuring and managing the externality of managerial responses to online customer reviews," *Information Systems Research*, 30(1), pp. 81-96.

Chen, Y., Ma, Y., Mao, Y., and Li., Q. 2019. "Multi-Task Learning for Abstractive and Extractive Summarization," *Data Science and Engineering*, 4(1), pp. 14-23.

Chen, Z., Trabelsi, M., Heflin, J., Xu, Y., and Davison, B.D. 2020. "Table Search Using a Deep Contextualized Language Model," *Proceedings of 43rd International ACM SIGIR Conference on Research and Development in Information Retrieval*, pp. 589–598.

Cho, H., Knijnenburg, B., Kobsa, A., and Li, Y. 2018. "Collective Privacy Management in Social Media: A Cross-Cultural Validation," *ACM Transactions on Computer-Human Interaction* 25(3), pp. 1-33.

Choromanski, K., Likhosherstov, V., Dohan, D., Song, X., Gane, A., Sarlos, T., Hawkins, P., Davis, J., Mohiuddin, A., Kaiser, L., and Belanger, D. 2021. "Rethinking Attention With Performers." *Proceedings of ICLR 2021.*

Chung, J., Gulcehre, C., Cho, K. and Bengio, Y., 2014. Empirical evaluation of gated recurrent neural networks on sequence modeling. *NIPS 2014 Workshop on Deep Learning.*

Chung, S., Animesh, A., Han, K., and Pinsonneault, A. 2020. "Financial returns to firms' communication actions on firm-initiated social media: evidence from Facebook business pages," *Information Systems Research*, 31(1), pp.258-285.

Clark, P., Etzioni, O., Khashabi, D., Khot, T., Mishra, B.D., Richardson, K., Sabharwal, A., Schoenick, C., Tafjord, O., Tandon, N., and Bhakthavatsalam, S. 2019. "From 'F' to 'A' on the NY Regents Science Exams: An Overview of the Aristo Project." Allen AI. (*arXiv preprint arXiv:1909.01958*).

Collobert, R., Weston, J., Bottou, L., Karlen, M., Kavukcuoglu, K., and Kuksa, P. 2011. "Natural language processing (almost) from scratch," *Journal of machine learning research*, 12, pp.2493-2537.

Collobert, R. and Weston, J. 2008. "A unified architecture for natural language processing: Deep neural networks with multitask learning," *Proceedings of ICML 2008*, pp. 160-167.

Conneau, A., Khandelwal, K., Goyal, N., Chaudhary, V., Wenzek, G., Guzmán, F., Grave, E., Ott, M., Zettlemoyer, L., and Stoyanov, V. 2019. "Unsupervised cross-lingual representation learning at scale," *Proceedings of ACL 2020*, pp. 8440-8451.

Cortés, K., Duchin, R., and Sosyura, D. 2016. "Clouded judgment: The role of sentiment in credit origination." *Journal of Financial Economics*, 121(2), pp.392-413.

Dai, Z., Yang, Z., Yang, Y., Carbonell, J., Le, Q.V., and Salakhutdinov, R. 2019. "Transformer-xl: Attentive Language Models Beyond a Fixed-Length Context," *Proceedings of ACL 2019*, pp. 2978–2988

Deng, S., Huang, Z., Sinha, A., and Zhao, H. 2018. "The Interaction Between Microblog Sentiment and Stock Return: An Empirical Examination," *MIS Quarterly*, 42(3), pp. 895-918.

de Vries, H., Bahdanau, D., and Manning, C. 2020. "Towards Ecologically Valid Research on Language User Interfaces," (*arXiv preprint*

*arXiv:2007.14435*).

Devlin, J., Chang, M.W., Lee, K., and Toutanova, K. 2018. "Bert: Pre-Training of Deep Bidirectional Transformers for Language Understanding," *Proceedings of NAACL 2019*, pp. 4171-4186.

Dong, W., Liao, S., and Zhang, Z. 2018. "Leveraging Financial Social Media Data for Corporate Fraud Detection," *Journal of Management Information Systems*, 35(2), pp.461-487.

Duan, X., Yu, H., Mingming, Y., Zhang, M., Luo, W., and Zhang, Y. 2019. "Contrastive Attention Mechanism for Abstractive Sentence Summarization," *Proceedings of EMNLP 2019*, pp. 3044-3053.

Fountaine, T., McCarthy, B., and Saleh, T. 2019. "Building the AI-Powered Organization," *Harvard Business Review*, 97(4), pp.62-73.

George, J., Gupta, M., Giordano, G., Mills, A., Tennant, V., and Lewis, C. 2018. "The Effects of Communication Media and Culture on Deception Detection Accuracy," *MIS Quarterly*, 42(2), pp. 551-575.

Ghiassi, M., Zimbra, D., and Lee, S., 2016. "Targeted Twitter Sentiment Analysis for Brands Using Supervised Feature Engineering and the Dynamic Architecture for Artificial Neural Networks," *Journal of Management Information Systems*, 33(4), pp.1034-1058.

Ghosal, D., Majumder, N., Gelbukh, A., Mihalcea, R., and Poria, S. 2020. "COSMIC: COmmonSense knowledge for eMotion Identification in Conversations," (*arXiv preprint arXiv:2010.02795*).

Gong, J., Abhishek, V. and Li, B., 2017. Examining the Impact of Keyword Ambiguity on Search Advertising Performance: A Topic Model Approach," *MIS Quarterly* 43(3), pp. 805-829.

Gonzalez, G. H., Tahsin, T., Goodale, B. C., Greene, A. C., and Greene, C. S. 2016. "Recent advances and emerging applications in text and data mining for biomedical discovery." Briefings in Bioinformatics, 17(1), pp. 33-42.

Gruetzemacher, R., Paradice, D., and Lee, K.B. 2020. "Forecasting extreme labor displacement: A survey of AI practitioners," *Technological Forecasting and Social Change*, 161.

Gu, Y., and Hu, Y. 2019. "Extractive Summarization with Very Deep Pretrained Language Model," *International Journal of Artificial Intelligence and Applications*, 10(2), pp. 27-32.

Gyawali, B., Anastasiou, L., and Knoth, P. 2020. "Deduplication of Scholarly Documents using Locality Sensitive Hashing and Word Embeddings," *Proceedings of 12th Language Resources and Evaluation Conference*, pp. 901-910.

Hao, Jie, Wang, X., Shi, S., Zhang, J., and Tu, Z. 2019. "Multi-Granularity Self-Attention for Neural Machine Translation," *Proceedings of EMNLP 2019*, pp. 887-897.

He, P., Liu, X., Gao, J. and Chen, W., 2021. "Deberta: Decoding-enhanced bert with disentangled attention." International Conference on Learning Representations.

Heimbach, I., and Hinz, O. 2018. "The impact of sharing mechanism design on content sharing in online social networks," *Information Systems Research*, 29(3), pp. 592-611.

Hochreiter, S., and Schmidhuber, J. 1997. "Long short-term memory," *Neural Computation*, 9(8), pp. 1735-1780.

Houlsby, N., Giurgiu, A., Jastrzebski, S., Morrone, B., DeLaroussilhe, Q., Gesmundo, A., Attariyan, M., and Gelly, S. 2019. "Parameter-efficient transfer learning for NLP," *Proceedings of the International Conference on Machine Learning*, pp. 2790-2799.

Howard, J., and Ruder, S. 2018. "Universal language model fine-tuning for text classification," (*arXiv preprint arXiv:1801.06146*).

Hoyle, A., Goel. P., and Resnik, P. 2020. "Improving Neural Topic Models using Knowledge Distillation," *Proceedings of EMNLP 2020*, pp. 1752-1771.

Hu, J., Ruder, S., Siddhant, A., Neubig, G., Firat, O., and Johnson, M. 2020. "Xtreme: A massively multilingual multi-task benchmark for evaluating cross-lingual generalization." *Proceedings of ICML 2020*, pp. 4411-4421.

Huang, J, Meng, Y., Guo, F., Ji, H., and Han, J. 2020. "Weakly-supervised aspect-based sentiment analysis via joint aspect-sentiment topic embedding," (*arXiv preprint arXiv:2010.06705*).

Huang, J., Boh, W., and Goh, K. 2017. "A temporal study of the effects of online opinions: Inform-ation sources matter," *Journal of Management Information Systems*, 34(4), pp. 1169-1202.

Huang, K. Y., Chengalur-Smith, I., and Pinsonneault, A. 2019. "Sharing is Caring: Social Support Provision and Companionship Activities in Healthcare Virtual Support Communities," *MIS Quarterly*, 43(2), pp. 395-424.

Huang, N., Hong, Y., and Burtch, G. 2017. "Social network integration and user content generation: Evidence from natural experiments," *MIS Quarterly*, 41(4), pp. 1035-1058.

Huang, P., Zhang, H., Jiang, R., Stanforth, R., Welbl, J., Rae, J., Maini, V., Yogatama, D., and Kohli, P. 2019. "Reducing sentiment bias in language models via counterfactual evaluation," *Proceedings of EMNLP 2019*, pp. 65-83.

Hwang, E. H., Singh, P. V., and Argote, L. 2019. "Jack of all, master of some: Information network and innovation in crowdsourcing communities," *Information Systems Research*, 30(2) pp. 389-410.

Jiang, F., Lee, J., Martin, X., and Zhou, G. 2019. "Manager Sentiment and Stock Returns," *Journal of Financial Economics*, 132(1), pp. 126-149.

Jin, H., Wang, T., and Wan, X. 2020. "Multi-Granularity Interaction Network for Extractive and Abstractive Multi-Document Summarization," *Proceedings of ACL 2020*, pp. 6244-6254.

Kaplan, J., McCandlish, S., Henighan, T., Brown, T., Chess, B., Child, R., Gray, S., Radford, A., Wu, J., Amodei, D. 2020. *Scaling laws for neural language models*, OpenAI. (arXiv preprint arXiv:2001.08361).

Ke, P., Ji, H., Liu, S., Zhu, X., and Huang, M. 2020. "SentiLARE: Linguistic Knowledge Enhanced Language Representation for Sentiment Analysis," *Proceedings of EMNLP 2020*, pp. 6975-6988.

Keskar, N. S., McCann, B., Varshney, L. R., Xiong, C., and Socher, R. 2019. "CTRL: A Conditional Transformer Language Model for Controllable Generation," (arXiv preprint arXiv:1909.05858).

Khashabi, D., Khot, T., Sabharwal, A., Tafjord, O., Clark, P., and Hajishirzi, H. 2020. "UnifiedQA: Crossing Format Boundaries With a Single QA System," *Proceedings of EMNLP 2020*, pp. 1896-1907.

Khern-am-nuai, W, Kannan, K., and Ghasemkhani, H. 2018. "Extrinsic Versus Intrinsic Rewards for Contributing Reviews in an Online Platform," *Information Systems Research*, 29(4), pp. 871-892.

Kitaev, N., Kaiser, Ł., and Levskaya, A. 2020. "Reformer: The Efficient Transformer." *Proceedings of ICLR 2020*.

Kraus, M., and Feuerriegel, S. 2017. "Decision support from financial disclosures with deep neural networks and transfer learning," *Decision Support Systems*, 104, pp. 38-48.

Kraus, M., Feuerriegel, S., and Oztekin, A. 2020. "Deep learning in business analytics and operations research: Models, applications and managerial implications," *European Journal of Operational Research*, 281(3), pp. 628-641.

Kryściński, W., Keshkar, N. S., McCann, B., Xiong, C., and Socher, R. 2019. "Neural text summarization: A critical evaluation," *Proceedings of EMNLP 2019*, pp. 540-551.

Kumar, N., Venugopal, D., Qiu, L., and Kumar, S. 2019. "Detecting Anomalous Online Reviewers: an Unsupervised Approach Using Mixture Models," *Journal of Management Information Systems*, 36(4), pp. 1313-1346.

Lappas, T., Sabnis, G., and Valkanas, G. 2016. "The Impact of Fake Reviews on Online Visibility: A Vulnerability Assessment of the Hotel Industry," *Information Systems Research*, 27(4), pp. 940-961.

Lan, Z., Chen, M., Goodman, S., Gimpel, K., Sharma., P., and Soricut, R. 2019. "ALBERT: A Lite BERT for Self-Supervised Learning of Language Representations." *Proceedings of ICLR 2019.*

LeCun, Y., Bengio, Y. and Hinton, G., 2015. "Deep Learning." *Nature*, (521), pp.436-444.

Lee, G.M., He, S., Lee, J., and Whinston, A.B. 2020. "Matching Mobile Applications for Cross-Promotion." *Information Systems Research*, 31(3), pp.865-891.

Lewis, M., Liu, Y., Goyal, N., Ghazvininejad, M., Mohamed, A., Levy, O., Stoyanov, V. and Zettlemoyer, L., 2020a. "Bart: Denoising sequence-to-sequence pre-training for natural language generation, translation, and comprehension." *Proceedings of ACL 2020.*

Lewis, M., Ghazvininejad, M., Ghosh, G., Aghajanyan, A., Wang, S. and Zettlemoyer, L., 2020b. "Pre-training via paraphrasing." *Advances in Neural information Processing Systems.*

Lewis, P., Perez, E., Piktus, A., Petroni, F., Karpukhin, V., Goyal, N., Küttler, H., Lewis, M., Yih, W.T., Rocktäschel, T. and Riedel, S., 2020c. Retrieval-augmented generation for knowledge-intensive nlp tasks. *Advances in Neural Information Processing Systems.*

Li, J., Larsen, K., and Abbasi, A. 2020. "TheoryOn: A Design Framework and System for Unlocking Behavioral Knowledge Through Ontology Learning," *MIS Quarterly*, 44(4), pp. 1733-1772.

Li, M., et al. "Applying BERT to Analyze Investor Sentiment in Stock Market," *Neural Computing and Applications*.

Li, W., Chen, H., and Nunamaker, J. F. 2016. "Identifying and Profiling Key Sellers in Cyber Carding Community: AZSecure Text Mining System," *Journal of Management Information Systems*, 33(4), pp. 1059-1086.

Li, X., Bing, L., Zhang, W., and Lam, W. 2019. "Exploiting BERT for End-to-End Aspect-Based Sentiment Analysis," *Proceedings of 2019 EMNLP Workshop W-NUT*, pp. 34-41.

Liang, H., Xue, Y., Pinsonneault, A., and Wu, Y. 2019. "What Users Do Besides Problem-Focused Coping When Facing IT Security Threats: An Emotion-Focused Coping Perspective," *MIS Quarterly*, 43(2), pp. 373-394.

Liang, Y., Duan, N., Gong, Y., Wu, N., Guo, F., Qi, W., Gong, M., Shou, L., Jiang, D., Cao, G., Fan, X., Zhang, R., Agrawal, R., Cui, E., Wei, S., Bhiarti, T., Qiao, Y., Chen, J.-H., Wu., W., Liu, S., Yang, F., Campos, D., Majumder, R., and Zhou, M. 2020. "XGLUE: A New Benchmark Dataset for Cross-Lingual Pre-Training, Understanding and Generation," *Proceedings of EMNLP 2020*, pp. 6008-6018.

Lieber, O., Sharir, O., Lenz, B. and Shoham, Y. 2021. "Jurrasic-1: Technical Details and Evaluation." White Paper, AI21 Labs.

Liu, X., Zhang, B., Susarla, A., and Padman, R. 2020a. "Go to YouTube and Call Me in the Morning: Use of Social Media for Chronic Conditions," *MIS Quarterly*, 44(1b), pp. 257-283..

Liu, X., He, P., Chen, W., and Gao, J. 2019a. "Multi-Task Deep Neural Networks for Natural Language Understanding," *Proceedings of ACL 2019*, pp. 4487-4496.

Liu, X., Wang, G. A., Fan, W., and Zhang, Z. 2020b. "Finding Useful Solutions in Online Knowledge Communities: A Theory-Driven Design and Multilevel Analysis," *Information Systems Research*, 31(3).

Liu, Y., and Lapata, M. 2019. "Text Summarization With Pretrained Encoders," Proceedings of EMNLP 2019b, pp. 3730-3740.

Liu, Y., Ott, M., Goyal, N., Du, J., Joshi, M., Chen, D., Levy, O., Lewis, M., Zettlemoyer, L. and Stoyanov, V. 2019c. "RoBERTa: A Robustly Optimized BERT Pretraining Approach," (arXiv preprint arXiv:1907.11692).

Liu, Y., Pant, G., and Sheng, O. R. L. 2020c. "Predicting Labor Market Competition: Leveraging Interfirm Network and Employee Skills," *Information Systems Research, forthcoming.*

Lu, Y., Dong, Y., and Charlin, L. 2020. "Multi-XScience: A Large-scale Dataset for Extreme Multi-document Summarization of Scientific Articles." *Proceedings of EMNLP 2020*, pp. 8068-8074.

Ma, X., Sap, M., Rashkin, H., and Choi, Y. 2020. "PowerTransformer: Unsupervised Controllable Revision for Biased Language Correction." *Proceedings of EMNLP 2020*, pp. 7426-7441.

Mai, F., Shan, Z., Bai, Q., Wang, X., and Chiang, R. H. L. 2018. "How does social media impact Bitcoin value? A Test of the Silent Majority Hypothesis," *Journal of Management Information Systems*, 35(1), pp. 19-52.

Manning, C., and Schütze, H. 1999. *Foundations of Statistical Natural Language Processing*, Cambridge, Massachusetts: MIT Press.

Maynez, J., Narayan, S., Bohnet, B., and McDonald, R. 2020. "On Faithfulness and Factuality in Abstractive Summarization," *Proceedings of EMNLP 2020*, pp. 1906-1919.

McCann, B., Bradbury, J., Xiong, C., and Socher, R. 2017. "Learned in Translation: Contextualized Word Vectors," *Advances in Neural Information Processing Systems.*

Mejia, J., Mankad, S., and Gopal, A. 2019. "A for Effort? Using the Crowd to Identify Moral Hazard in New York City Restaurant Hygiene Inspections," *Information Systems Research*, 30(4), pp. 1363-1386.

Mikolov, T., Sutskever, I., Chen, K., Corrado, G.S. and Dean, J., 2013a. Distributed representations of words and phrases and their compositionality. In *Advances in neural information processing systems* (pp. 3111-3119).

Mikolov, T., Chen, K., Corrado, G. S., and Dean, J. 2013b. "Efficient Estimation of Word Representations in Vector Space," *Proceedings of ICLR 2013.*

Mikolov, T., Joulin, A., Chopra, S, Mathieu, M., and Ranzato, M.A. 2014. "Learning Longer Memory in Recurrent Neural Networks," (*arXiv preprint arXiv:1412.7753*).

Moreno, A., and Redondo, T. 2016. "Text Analytics: The Convergence of Big Data and Artificial Intelligence," *International Journal of Interactive Multimedia and Artificial Intelligence*, 3(6), pp. 57-64.

Frank, M. R., Autor, D. Bessen, J. E., Brynjolfsson, E., Cebrian, M., Deming, D. J., Feldman, M., Groh, M., Lobo, J., Moro, E., and Wang, D. 2019. "Toward Understanding the Impact of Artificial Intelligence on Labor," *Proceedings of the National Academy of Sciences*, 116(14), pp. 6531-6539.

Mousavi, R., and Gu, B. 2019. "The Impact of Twitter Adoption on Lawmakers' Voting Orientations," *Information Systems Research*, 30(1), pp. 133-153.

Mousavi, R., Johar, M., and Mookerjee, V. J. "The Voice of the Customer: Managing Customer Care in Twitter," *Information Systems Research*, 31(2), pp. 340-360.

Müller, M., Salathé, M., and Kummervold, P. E. 2020. "COVID-Twitter-BERT: A Natural Language Processing Model to Analyse COVID-19 Content on Twitter," (*arXiv preprint arXiv:2005.07503*).

Murphy, K. P. 2012. *Machine Learning: A Probabilistic Perspective*. Cambridge, Massachusetts: MIT press.

Narayan, S., Cohen, S. B., and Lapata, M. 2019. "What is this Article about? Extreme Summarization with Topic-aware Convolutional Neural Networks," *Journal of Artificial Intelligence Research*, 66, pp. 243-278.

Naseem, U., Razzak, I., Musial, K. and Imran, M. 2020. "Transformer based deep intelligent contextual embedding for twitter sentiment analysis," *Future Generation Computer Systems*, 113, pp. 58-69.

Ngai, E. W. T., and Lee, P. T. Y. 2016. "A Review of the Literature on Applications of Text Mining in Policy Making," *Proceedings of PACIS 2016.*

Nguyen, D. Q., Vu, T., and Nguyen, A. T.. "BERTweet: A Pre-Trained Language Model for English Tweets," *Proceedings of EMNLP 2020: Demonstrations*, pp. 9-14.

Pan, Y., Huang, P., and Gopal, A. 2019. "Storm Clouds on the Horizon? New Entry Threats and R&D Investments in the US IT Industry," *Information Systems Research*, 30(2), pp. 540-562.

Park, J., Park, C., Kim, J., Cho, M., and Park, S. 2019. "ADC: Advanced Document Clustering using Contextualized Representations," *Expert Systems with Applications*, 137, pp. 157-166.

Pennington, J., Socher, R., and Manning, C. 2014. "Glove: Global vectors for word representation." *Proceedings of EMNLP 2014*, pp. 1532-1543.

Peters, M., Neumann, M., Iyyer, M., Gardner, M, Clark, C., Lee, K., and Zettlemoyer, L. 2018 "Deep contextualized word representations." *Proceedings of NAACL 2018*, pp. 2227-2237.

Radford, A., Narasimhan, K., Salimans, T., and Sutskever, I. 2018. *Improving Language Understanding by Generative Pre-Training*, OpenAI.

Radford, A., Wu, J., Child, R., Luan, D., Amodei, D., and Sutskever, I. 2019. *Language Models are Unsupervised Multitask Learners*, OpenAI.

Raffel, C., Shazeer, N., Roberts, A., Lee, K., Narang, S., Matena, M., Zhou, Y., Li, W. and Liu, P.J.,2020. "Exploring the Limits of Transfer Learning with a Unified Text-to-Text Transformer," *Journal of Machine Learning Research*, 21(140):1−67.

Reimers, N., Schiller, B., Beck, T., Daxenberger, J., Stab, C., and Gurevych, I. 2019. "Classification and Clustering of Arguments with Contextualized Word Embeddings," *Proceedings of ACL 2019*, pp. 567-578

Rhue, L., and Sundararajan, A. 2019. "Playing to the Crowd? Digital Visibility and the Social Dynamics of Purchase Disclosure," *MIS Quarterly*, 43(4), pp. 1127-1141.

Rocklage, M. D., and Fazio, R. H. "The Enhancing Versus Backfiring Effects of Positive Emotion in Consumer Reviews," *Journal of Marketing Research*, 57(2), pp. 332-352.

Roller, S., Dinan, E., Goyal, N., Ju, D., Williamson, M., Liu, Y., Xu, J., Ott, M., Shuster, K., Smith, E.M., and Boureau, Y.L. 2020a. "Recipes for Building an Open-Domain Chatbot," (*arXiv preprint arXiv:2004.13637*).

Roller, S., Boureau, Y.L., Weston, J., Bordes, A., Dinan, E., Fan, A., Gunning, D., Ju, D., Li, M., Poff, S., and Ringshia, P. 2020b. "Open-Domain Conversational Agents: Current Progress, Open Problems, and Future Directions," (*arXiv preprint arXiv:2006.12442*).

Saifee, D. H., Bardhan, I.R., Lahiri, A., and Zheng, Z. 2019. "Adherence to Clinical Guidelines, Electronic Health Record Use, and Online Reviews," *Journal of Management Information Systems* 36(4), pp. 1071-1104.

Salant, S., and Berant, J. 2018. "Contextualized Word Representations for Reading Comprehension," *Proceedings of NAACL 2018*, pp. 554-

559.

Salo, M., Mykkänen, M., and Hekkala, R. 2020. "The Interplay of IT Users' Coping Strategies: Uncovering Momentary Emotional Load, Routes, and Sequences," *MIS Quarterly*, 44(3), pp. 1143-1175.

Samtani, S., Chinn, R, Chen, H., and Nunamaker, J. F. 2017. "Exploring emerging hacker assets and key hackers for proactive cyber threat intelligence," *Journal of Management Information Systems*, 34(4), pp. 1023-1053.

Samtani, S., Zhu, H., Padmanabhan, B., Chai, Y., and Chen, H. 2020. "Deep Learning for Information Systems Research," (*arXiv preprint arXiv:2010.05774*).

Sanh, V., Debut, L., Chaumond, J., and Wolf, T. 2019. "DistilBERT, a Distilled Version of BERT: Smaller, Faster, Cheaper and Lighter," *NeurIPS Workshop on Energy Efficient Machine Learning and Cognitive Computing*.

Schick, T., and Schütze, H. 2020. "It's Not Just Size That Matters: Small Language Models Are Also Few-Shot Learners," (*arXiv preprint arXiv:2009.07118*).

Schwartz, D., Toneva, M., and Wehbe, L. 2019. "Inducing brain-relevant bias in natural language processing models," *Advances in Neural Information Processing Systems*.

Scialom, T., Dray, P.-A., Lamprier, S., Piwowarski, B., and Staiano, J. 2020. "MLSUM: The Multilingual Summarization Corpus." *Proceedings of EMNLP 2020*, pp. 8051-8067.

Shi, D., Guan, J., Zurada, J., and Manikas, A. 2017. "A Data-Mining Approach to Identification of Risk Factors in Safety Management Systems," *Journal of Management Information Systems*, 34(4), pp. 1054-1081.

Shi, Z., Lee, G. M., and Whinston, A. B. 2016. "Toward a Better Measure of Business Proximity: Topic Modeling for Industry Intelligence," *MIS Quarterly*, 40(4), pp. 1035-1056.

Shin, D., He, S., Lee, G.M., Whinston, A.B., Cetintas, S. and Lee, K.C. 2019. "Enhancing Social Media Analysis with Visual Data Analytics: A Deep Learning Approach," *MIS Quarterly*, 2020.

Sia, S., Dalmia, A., and Mielke, S. J. "Tired of Topic Models? Clusters of Pretrained Word Embedd-ings Make for Fast and Good Topics too!," *Proceedings of EMNLP 2020*, pp. 1728-1736.

Siering, M., Koch, J. and Deokar, A. 2016. "Detecting Fraudulent Behavior on Crowdfunding Platforms: The Role of Linguistic and Content-Based Cues in Static and Dynamic Contexts," *Journal of Management Information Systems*, 33(2), pp. 421-455.

Silver, D., Schrittwieser, J., Simonyan, K., Antonoglou, I., Huang, A., Guez, A., Hubert, T., Baker, L., Lai, M., Bolton, A., and Chen, Y. 2017. "Mastering the Game of Go Without Human Knowledge," *Nature*, 550(7676), pp. 354-359.

Smith, N. A. 2020. "Contextual word representations: putting words into computers." *Communications of the ACM* 63(6), pp. 66-74.

Song, T., Huang, J., Tan, Y., and Yu, Y. 2019. "Using User-and Marketer-Generated Content for Box Office Revenue Prediction: Differences Between Microblogging and Third-Party Platforms," *Information Systems Research*, 30(1), pp. 191-203.

Stiennon, N., Ouyang, L., Wu, J., Ziegler, D., Lowe, R., Voss, C., Radford, A., Amodei, D., and Christiano, P.F. 2020. "Learning to Summarize with Human Feedback," *Advances in Neural Information Processing Systems*, 33.

Sun, C., Qiu, X., Xu, Y., and Huang, X. 2019. "How to Fine-Tune BERT for Text Classification?," *China National Conference on Chinese Computational Linguistics*, Springer, pp. 194-206.

Sun, Y., Wang, S., Feng, S., Ding, S., Pang, C., Shang, J., Liu, J., Chen, X., Zhao, Y., Lu, Y. and Liu, W., 2021. ERNIE 3.0: Large-scale Knowledge Enhanced Pre-training for Language Understanding and Generation. *arXiv preprint arXiv:2107.02137*.

Tan, H., and Bansal, M. 2019. "LXMERT: Learning Cross-Modality Encoder Representations from Transformers," *Proceedings of EMNLP*, pp. 5100-5111.

Tan, H., and Bansal, M. 2020. "Vokenization: Improving Language Understanding with Contextualized, Visual-Grounded Supervision," *Proceedings of EMNLP*, pp. 2066-2080.

Tay, Y., Dehghani, M., Gupta, J., Bahri, D., Aribandi, V., Qin, Z. and Metzler, D., 2021. Are Pre-trained Convolutions Better than Pre-trained Transformers?. *Proceedings of ACL 2021*.

Turing, A. M. 1950. "Computing Machinery and Intelligence," *Mind*, 59(236), pp. 433-460.

Van Osch, W., and Steinfield, C. W. 2018. "Strategic Visibility in Enterprise Social Media: Implications for Network Formation and Boundary Spanning," *Journal of Management Information Systems*, 35(2), pp. 647-682.

Vaswani, A., Chang, M.W., Lee, K., and Toutanova, K. 2017. "Attention is all You Need," *Advances in Neural Information Processing Systems*.

Velichety, S., Ram, S., and Bockstedt, J. "Quality Assessment of Peer-Produced Content in Knowledge Repositories Using Development and Coordination Activities," *Journal of Management Information Systems*, 36(2), pp. 478-512.

Wang, A., Singh, A., Michael, J., Hill, F., Levy, O., and Bowman, S.R. 2018a. "GLUE: A Multi-Task Benchmark and Analysis Platform for Natural Language Understanding," *EMNLP Workshop on BlackBox NLP*, pp 353-355.

Wang, A., Pruksachatkun, Y., Nangia, N., Singh, A., Michael, J., Hill, F., Levy, O., and Bowman, S. 2019a. "SuperGLUE: A Stickier Benchmark for General-Purpose Language Understanding Systems," *Advances in Neural Information Processing Systems*.

Wang, Q., Li, B., and Singh, P. V. 2018b. "Copycats vs. Original Mobile Apps: A Machine Learning Copycat-Detection Method and Empirical Analysis," *Information Systems Research*, 29(2), pp. 273-291.

Wang, X., and Liu, Z. 2019b. "Online Engagement in Social Media: A Cross-Cultural Comparison," *Computers in Human Behavior*, 97, pp. 137-150.

Wang, Z., Jiang, C., Zhao, H., and Ding, Y. 2020. "Mining Semantic Soft Factors for Credit Risk Evaluation in Peer-to-Peer Lending," *Journal of Management Information Systems*, 37(1), pp. 282-308.

Wang, Z., Tsvetkov, Y., Firat, O., and Cao, Y. 2021a. "Gradient Vaccine: Investigating and Improving Multi-task Optimization in Massively Multilingual Models," *Proceedings of ICLR 2021*.

Wang, Z., Yu, A.W., Firat, O. and Cao, Y., 2021b. Towards Zero-Label Language Learning. *arXiv preprint arXiv:2109.09193*.

Wei, J., Bosma, M., Zhao, V.Y., Guu, K., Yu, A.W., Lester, B., Du, N., Dai, A.M. and Le, Q.V., 2021. Finetuned language models are zero-shot learners. *arXiv preprint arXiv:2109.01652*.

Wolf, T., Chaumond, J., Debut, L., Sanh, V., Delangue, C., Moi, A., Cistac, P., Funtowicz, M., Davison, J., Shleifer, S., and Louf, R. 2020. "Transformers: State-of-the-Art Natural Language Processing," *Proceedings of EMNLP 2020: Demonstrations*, pp. 38-45.

Wu, C.S., Hoi, S., Socher, R., and Xiong, C. 2020. "TOD-BERT: Pre-Trained Natural Language Understanding for Task-Oriented Dialogues." *Proceedings of EMNLP 2020*, pp. 917-929.

Wu, J., Huang, L., and Zhao, J. L. 2019. "Operationalizing Regulatory Focus in the Digital Age: Evidence from an E-Commerce Context," *MIS Quarterly*, 43(3), pp. 745-764.

Wu, J., Ouyang, L., Ziegler, D.M., Stiennon, N., Lowe, R., Leike, J. and Christiano, P., 2021. Recursively summarizing books with human feedback. *arXiv preprint arXiv:2109.10862*.

Xia, P., Wu, S., and Van Durme, B. 2020. "Which* BERT? A Survey Organizing Contextualized Encoders," *Proceedings of EMNLP 2020*, pp. 7516-7533.

Xu, H., Liu, B., Shu, L., and Yu, P.S. 2019. "BERT Post-Training for Review Reading Compre-hension and Aspect-based Sentiment Analysis," *Proceedings of NAACL 2019*, pp. 2324-2335.

Xu, K., Ba, J., Kiros, R., Cho, K., Courville, A., Salakhudinov, R., Zemel, R. and Bengio, Y., 2015, June. Show, attend and tell: Neural image caption generation with visual attention. In *International conference on machine learning* (pp. 2048-2057). PMLR.

Xu, Y., and Lapata, M. 2020. "Coarse-to-Fine Query Focused Multi-Document Summarization," *Proceedings of EMNLP 2020*, pp. 3632-3645.

Yang, M., Adomavicius, G., Burtch, G., and Ren, Y. 2018. "Mind the Gap: Accounting for Measurement Error and Misclassification in Variables Generated via Data Mining," *Information Systems Research*, 29(1), pp. 4-24.

Yang, R., Cao, J., Wen, Z., Wu, Y., and He, X. 2020. "Enhancing Automated Essay Scoring Performance via Cohesion Measurement and Combination of Regression and Ranking," *Proceedings EMNLP 2020: Findings*, pp. 1560-1569.

Yang, Z., Dai, Z., Yang, Y., Carbonell, J., Salakhutdinov, R.R., and Le, Q.V. 2019. "XLNet: Generalized Autoregressive Pretraining for Language Understanding," *Advances in Neural Information Processing Systems*.

Yao, L., Jin, Z., Mao, C., Zhang, Y. and Luo, Y. 2019. "Traditional Chinese Medicine Clinical Records Classification with BERT and Domain Specific Corpora," *Journal of the American Medical Informatics Association*, 26(12), pp. 1632-1636.

Yoo, E., Gu, B., and Rabinovich, E. 2019. "Diffusion on Social Media Platforms: A Point Process Model for Interaction among Similar Content," *Journal of Management Information Systems*, 36(4), pp. 1105-1141.

Young, H. 2015. "The digital language divide," *The Guardian*, (*http://labs.theguardian.com/digital-language-divide/*).

Yue, W. T., Wang, Q.-H., and Hui, K.-L. 2019. "See no evil, hear no evil? Dissecting the impact of online hacker forums," *MIS Quarterly*, 43(1), pp. 73-95.

Zaheer, M., Guruganesh, G., Dubey, A., Ainslie, J., Alberti, C., Ontanon, S., Pham, P., Ravula, A., Wang, Q., Yang, L., and Ahmed, A. 2020. "Big Bird: Transformers for Longer Sequences," *Advances in Neural Information Processing Systems*, 33.

Zellers, R., Bisk, Y., Farhadi, A., and Choi, Y. 2019. "From recognition to cognition: Visual commonsense reasoning." *Proceedings of CVPR 2019*, pp. 6720-6731.

Zhang, D., Zhou, L., Kehoe, J. L., and Kilic, I. Y. et al. 2016. "What online reviewer behaviors really matter? Effects of verbal and nonverbal behaviors on detection of fake online reviews," *Journal of Management Information Systems*, 33(2), pp. 456-481.

Zhang, K., Bhattacharyya, S., and Ram, S. 2016. "Large-Scale Network Analysis for Online Social Brand Advertising," *MIS Quarterly*, 40(4), pp. 849-868.

Zhang, S., Zhao, H., Wu, Y., Zhang, Z., Zhou, X., and Zhou, X. 2020. "DCMN+: Dual Co-Matching Network for Multi-Choice Reading Comprehension," *Proceedings of AAAI 2020*, 34(5), pp. 9563-9570.

Zhang, W., and Ram, S. 2020. "A Comprehensive Analysis of Triggers and Risk Factors for Asthma Based on Machine Learning and Large Heterogeneous Data Sources," *MIS Quarterly*, 44(1), pp. 305-339.

Zhong, P., Wang, D., and Miao, C. 2019. "Knowledge-enriched transformer for emotion detection in textual conversations." *Proceedings of EMNLP 2019*, pp. 165-176.

Zhou, S., Qiao, Z., Du, Q., Wang, G.A., Fan, W. and Yan, X. 2018. "Measuring Customer Agility From Online Reviews Using Big Data Text Analytics." *Journal of Management Information Systems*, 35(2), pp. 510-539.

Zhu, J., Wang, Q., Wang, Y., Zhou, Y., Zhang, J., Wang, S., and Zong, C. 2019. "NCLS: Neural Cross-Lingual Summarization," *Proceedings of EMNLP 2019*, pp. 3054-3064.

# A  APPENDIX OF EXTENDED LITERATURE REVIEW TABLES

Tables A1-A3 depict summaries and recommendations for each item in the IS literature review.

## Table A1. Information Systems Research Survey Summaries and Recommendations

| Study | Summary | Recommendations (*e.g.*, future work) |
|---|---|---|
| Abbasi et al. 2019 | This study examines key design elements of Voice of the Customer (VoC) listening platforms via a novel and improved "mention model" that is designed for early detection of adverse events (to address limitations of existing work). The model is evaluated using tweets and other web content involving adverse events in the automotive and pharmaceutical industries. The entire model is built simply on mentions, and only utilizes text mining in combination with signal fusion to improve recall (SentiWordNet lexicon for sentiment analysis). | Simply using DTL would likely yield marginal to modest improvement for the sensor fusion model. However, significant improvement could be realized leveraging the gains offered by TLMs over the supervised and unsupervised machine learning methods that are used for studies cited in the review of methods. Because this is effectively a predictive model, it is possible that with a labeled training dataset the creation of a strong model for this task using TLM fine-tuning could become very easy to develop. In general, VoC listening seems like a research topic ripe for the application of TLMs in future work. |
| Adamopoulos et al. 2018 | This study uses machine learning, including several text mining techniques, to determine whether personality characteristics of social media users accentuate or attenuate the effectiveness of word of mouth in shaping consumers' preferences and behavior. The model uses latent Dirichlet allocation (LDA) for topic modeling and for generating a single feature for each topic as well as a commercial deep learning sentiment analysis package. | As the model relies heavily on traditional text mining techniques, and only uses an unknown deep learning technique for sentiment analysis, the overall gains the model could see from incorporating DTL would likely be moderate to significant. However, the amount of benefit is difficult to estimate given that LDA may not see more than moderate improvement and the deep learning technique used for sentiment analysis is unknown. Alternately, CWRs could be a better option for feature extraction. |
| Biohm et al. 2016 | This study explores differences in behavioral decision making as a result of IT-based support of open idea evaluation through a randomized experiment. Text mining is used to evaluate readability (through computation of frequency-based metrics of readability) in order to assess understandability as one variable in their model. | This study uses text mining in a minimal and unique manner. Thus, it is unlikely that there would be much improvement to the overall model if DTL or CWRs were used. (Readability is not a topic for which the latest techniques have been demonstrated, so we may underestimate the potential for improvement. It is possible that a methodological contribution could come from the development of a measure of readability using TLMs or CWRs.) |
| Chen et al. 2020 | This study uses a text analytics approach to extract network features from news for better understanding market signaling via dynamic firm relations. Specifically, text mining is used to extract dynamic firm networks from news via steps involving event extraction, argument separation and sentiment analysis. The results suggest new research directions for financial text mining and new insights into investors' perceptions of news. | The study describes existing text mining technologies as being severely limited with respect to being able to understand the semantics of the inter-firm relationships that the study focuses on. For this reason, and because the authors underscore the value of financial news for market signaling and other applications, we feel that there is the potential for significant improvement using either CWRs or DTL. (Krauss & Feuerriegel [2017] demonstrate the value of DTL, without utilizing TLMs or CWRs, for directly predicting firms' market movements, suggesting that significant gains are possible with greater ease due to substantially less model complexity.) |
| Chen et al. 2019 | This study assesses the magnitude of direct and indirect effects of managerial responses to online customer reviews. The study utilizes a difference-in-difference-in-differences approach and uses text mining to extract sentiment from reviews for use as a model feature. This is done with an n-gram language model and LingPipe software. | Because the study utilizes sentiment analysis as one component of its model, significant improvement cannot be expected with the use of DTL. However, its use could still be of marginal to moderate benefit. Because a commercial sentiment package is utilized, it is difficult to estimate any performance gains. |
| Chung et al. 2020 | This study evaluates the impact of firm social media messaging on firm performance (specifically on market returns). The study tests hypotheses regarding the impact on returns of firms' social media responses to positive and negative customer messages. It uses text mining for sentiment analysis so that sentiment scores could be used as features for customer comments, other social media content, blogs and news. Open Hangul, a foreign sentiment analysis package, was used for analysis. | Because this study uses sentiment analysis as a primary component in its model, the technique is very important. However, as the data involved firms' messages from three languages, this is not as straightforward as simply applying deep transfer learning or CWRs. Yet, this is a great example for how multilingual TLMs can be applied - due to the three languages different sentiment analysis techniques were required, but a single multilingual model could strengthen the model by ensuring that the sentiment values were learned from different languages and generated from the same semantic space. |
| Heimback & Hinz 2018 | This experimental study evaluates how design aspects of online social networks impact online content diffusion through a controlled experiment. The study only uses text mining in a minor capacity via sentiment analysis in order to enhance the dataset by adding extra features. Pennebaker et al.'s [2007] LIWC dictionary was used for sentiment analysis. | Because the study only uses sentiment analysis in a minor role for the model, and is primarily concerned with the experiment's results, it is unlikely that anything other than marginal benefit could come from the use of more advanced techniques such as DTL or CWRs. |
| Hwang et al. 2019 | The study examines how the information accumulated by individuals in a customer support crowdsourcing community impacts their ability to generate novel, feasible and well-liked ideas in an innovation crowdsourcing community. It uses NLP to construct each individual's information network based on their activities in the community segmented by topic (using topic modeling, i.e., LDA). The results indicate that generalists who have contributed to a broad range of topics in the online community are able to generate novel ideas more often than nongeneralists. | Since topic modeling is the method used in this study it is not clear how much benefit could be gained from using DTL or CWRs. While the use of these methods would likely be able to improve the results, the gains may be marginal at the cost of more effort (because topic modeling is not yet a widely used application of TLMs). However, advanced models that use CWRs for feature extraction stand to offer more substantive gains which may justify the effort. Regardless, it is unlikely that any changes would significantly impact the results of the study. |
| Khem-am-naui et al. 2018 | This study explores the implications of monetary incentives for reviews through a natural experiment that compared a review platform offering monetary incentives to reviews from Amazon. The study employs a variety of text mining techniques including a GF index (a metric computed via frequency analysis) for readability, sentiment analysis (used Harvard General Inquirer lexicon) and topic modeling (via LDA). These techniques all generated features to use with a difference-in-differences model. | Because text features are so prominent in the model used here, we feel that TLMs and CWRs are strong candidates for improving the quality of the model, especially due to biases introduced by uncertainty in feature extraction techniques [Yang 2018]. Again, we are not aware of TLMs being used for quantifying readability, but, as this appears to have substantial value to IS researchers, it may be that a methodological contribution developed from TLMs is useful to researchers and practitioners. We feel that since multiple features are extracted using text mining, the impact of using TLMs or CWRs would be significant, although it may not impact the results. |
| Lappas et al. 2016 | This study explores the detection of fake reviews on online platforms with respect to the vulnerability of firms to fake review attacks. To these ends the study focuses on the visibility of a business to the customer base as a function of the features that a business can cover and its position in the platform's ranking system. The authors employ a lexicon-based sentiment analysis aggregated at the hotel level as features for their model. | This work relies heavily on sentiment analysis, thus the method is quite relevant. In the past lexicon-based techniques have worked well for domain-specific sentiment analysis tasks because semantic distributions shift dramatically between different domains and often outperformed even alternative methods trained on domain-specific data. However, DTL using TLMs will outperform even these lexicon based methods with a reasonable training dataset. Performance gains may not be dramatic for models such as a pretrained BERT, but with larger datasets domain-specific pretraining could enable stronger models with fine-tuning from small datasets. Few-shot learning models could also be expected to do well. |
| Lee et al. 2020 | This study introduces cross-promotion - advertising one mobile app in another mobile app - as a new means for advertising apps. The authors create a novel app similarity measure by using LDA on apps' production descriptions, and then analyze how the similarity between the source app and the target app influences users' app downloads. The dataset is unique, and comes from a major mobile app provider in South Korea. They further propose a predictive machine learning app-matching system. | In this study, LDA was selected for topic modeling rather than a word representation model because of its wide acceptance in research contexts and because it was more interpretable. While there may be some truth to these claims, the performance benefits from CWRs likely outweigh each, especially considering that interpretation of text clustering is significantly easier than interpretation of high dimensional clusters of numeric data. Consequently, we feel that research like this would benefit from the use of TLM-based methods. |
| **Liu et al. 2020** | This study proposes a novel text analytic framework developed using design science for extracting important features from online collaborative forums and applying them to classify the usefulness of a solution. A kernel theory of the knowledge adoption model is used to capture a rich set of features for predicting information usefulness. n-gram vectorization was used with a variety of other techniques, such as topic modeling, to extract features for the classification model (several ML methods were used for the classification). (*See Figure 3 in paper*). | Since feature extraction plays such a big role in this study, it is likely that CWRs and TLMs would significantly boost performance. Because this is a predictive model, performance gains are more significant than for models that are used for statistical analysis. Thus, we feel that this study is a good example of one in which TLM-based methods would be very useful. This model is also a good candidate for using higher dimensional latent feature vectors, particularly because SVMs are used as the classifier and are well-suited for large feature spaces when there is sufficient data. We feel that using contextual word representations in this manner could dramatically improve predictive ability. |
| **Liu et al. 2020 (forthcoming)** | This study conducts an interfirm labor market competitor analysis with a longitudinal dataset derived from online profiles and matched by employer-employee to extract global information about interfirm human capital competition. Using employee transfer between firms they derive and analyze a human capital flow network. LDA is used for extracting a feature based on six topics. Other features are created by computing cosine similarity between descriptions (e.g. documents) using a technique similar to TF-IDF for vectorization. These features are then used as inputs for a variety of machine learning predictive models. | We anticipate that substantial improvement on these techniques is possible if DTL was used for the similarity feature extraction as opposed to simply using the TF-IDF derived technique. Due to the significance of context, and its strength relative to frequency-based methods, we are particularly interested in how significant improvement could be from their use, particularly because this is a predictive model where improved performance is the objective. Moreover, it is likely that other features could be extracted using CWRs, and that feature extraction via topic modeling could be enhanced to improve performance through their use. |
| **Mejia et al. 2019** | This study attempts to use information from online reviews to gauge restaurant hygiene. To these ends the study constructs a social media hygiene (SMASH) dictionary from the online reviews drawing on previous work utilizing sentiment and opinion mining. Mturk was used to classify each review based on whether it "indicates problems related to hygiene." This resulted in 1,191 training documents. 1-grams, 2-grams and 3-grams were included in the dictionary, which was generated using word frequency. | Because this study uses machine learning to generate a dictionary which is then used for analysis, it is likely that the unique training dataset used could be effectively used to fine-tune a pretrained TLM to achieve strong performance for the task of evaluating restaurant hygiene through sentiment/opinion analysis. An even stronger model could certainly be created by pretraining a TLM on a domain-specific dataset of restaurant reviews in an unsupervised fashion, and then fine-tuning with the crowd-generated labeled dataset. We also think that few-shot learning models could perform well on this task, although how well is unclear and they may need more than single digit numbers of examples. |
| **Mousavi & Gu 2019** | This study assessed the impact of US Representatives' Twitter adoption on their voting orientations in the US Congress, specifically, whether or not their adoption of Twitter makes them vote more in line with their constituents. Sentiment was used as a feature for the model by averaging four separate sentiment scores on Twitter data. | Marginal gains may be realized by using DTL to fine-tune a pretrained TLM on Twitter data. Perhaps even more significant gains could come from using a domain-specific pretrained TLM for fine-tuning further with a labeled dataset. This latter technique seems particularly well suited for Twitter data due to the syntactic irregularities as well as the language used. It could further be possible to encode emojis in text and pretrain a TLM which had tokenized them for social media data where they are widely used in ways that may lead to poor performance with existing techniques. However, in this particular study these techniques are unlikely to improve the results as sentiment is just one feature. |
| **Mousavi et al. 2020** | This study strives to identify an optimal strategy for managing customer sentiment social media and to identify factors influencing the effectiveness of managing customer sentiment in this manner. Consequently, sentiment analysis is used to generate a sentiment score for each Tweet via the lexicon-based R package SentimentR and the resulting sentiment scores were then normalized. This was then used as a primary feature for the model. | As with the Mousavi and Gu [2019] study, this study involved sentiment analysis of Twitter data. However, unlike the study above, sentiment was the primary feature for the model. Thus, this study stands to gain more from the use of DTL for improving the quality of sentiment analysis. While existing sentiment analysis techniques may perform better for many narrow domains, the use of TLMs pretrained on domain-specific datasets offers an avenue for improving upon the standard techniques used for IS research. Thus, work on methodological contributions along these lines would be welcome as it would stand to improve the quality of insights from research such as this which relies heavily on sentiment analysis. |

| Study | Summary | Recommendations (*e.g.*, future work) |
|---|---|---|
| Pan et al. 2019 | This study develops and validates a measure of threats posed by the entry of new startups in the IT industry using text mining to analyze 10-K filings. The authors create a novel measure of new entry threats that builds on TF-IDF vectorization and the cosine similarity measure. | This is another case that stands to benefit significantly from the use of TLMs. Rather than computing document similarity, DTL with TLMs can be used to perform extractive or abstractive summarization followed by single sentence similarity analysis, which TLMs perform quite well at based on their performance on these tasks in GLUE [Wang 2018]. Applying similar techniques in such a manner would be challenging but could lead to a superior novel measure for evaluating new entry threats. Again, like many applications discussed here, pretraining a TLM on a domain-specific corpus and then fine-tuning on a labeled dataset would have a significant positive impact on performance. |
| Song et al. 2019 | This study involves the development of a prediction model for box office revenue based on user-generated content and marketer-generated content from a popular Chinese microblogging service. The authors use a small portion of the dataset to label for training a sentiment classifier and they use a Chinese word segmentation repository for extracting features. A sentiment feature and the latent extracted features are fed to an SVM classifier which is then used as the sentiment. | Because the sentiment analysis used in this study has been labeled for training a sentiment model, it is a good candidate for applying DTL. Furthermore, because the content is all Chinese, it is a good candidate for use with a multilingual TLM or a TLM pretrained in Chinese. Due to the power of TLMs as predictive models and classifiers, it would be interesting to evaluate the performance differences that result from 1) using CWRs and deep transfer learning for feature extraction and use with an SVM and 2) training the model directly to predict box office revenue. This is a unique foreign language model and it would be interesting to explore the gains via either method. |
| Wang et al. 2018 | This study uses a combination of image processing, text mining and network analytics in a novel method to identify apps as original or copycat. For the text component LSA is used for feature generation and used for clustering to create a predictive model. The results are used in an econometric model to evaluate the impact of copycat apps on demand for the original apps. | Given the importance of feature extraction in this predictive model, it is likely that improvements can be realized by utilizing CWRs. However, it is more difficult to anticipate the significance of such gains from this model since it uses such a diverse range of machine learning techniques in combination. Thus, while we feel that the model would benefit from using TLM-based methods, we also feel that it may not benefit as much as other models discussed here. |
| Yang et al. 2018 | This study examines the nature of bias in models that use machine learning for extracting features to input into analytical models. This bias arises from measurement error or from the misclassification of variables. For one of the examples, the authors trained a text classifier on online reviews and for another example they trained a text classifier on social media posts. A third example involved a more traditional dataset involving structured numerical data. The results have interesting implications for future work utilizing machine learning techniques for feature extraction, and suggest that the best results will be achieved by coupling the best performing techniques with error-correction methods. | This study is particularly relevant to the present study as it explores the significance of error propagation from methods used for feature extraction to the analytical model. We feel strongly that the bias identified in the two text mining examples demonstrated here could be mitigated substantially via the use of DTL and CWRs. Moreover, we feel that this article demonstrates one of the significant reasons for adopting the use of these methods more widely in IS research. |

Table A2: The Journal of MIS Survey Summaries and Recommendations

| Study | Summary | Recommendations (*e.g.*, future work) |
|---|---|---|
| Chen et al. 2019 | This study addresses how consumer-to-consumer online health community users are affected by the relationships established in the communities and how the content exchanged between users impacts their health knowledge and attitudes. The study examines these questions using a computational multi-method framework that combines NLP, machine learning, social network analysis and econometric modeling to a panel dataset collected from nine separate discussion boards. The authors used POS tagging and LDA as features for an SVM classifier to sort posts. They then used SentiStrength to extract sentiment features from community posts (scaled -5 to 5) for input into an econometric model. | This study utilizes a range of text mining/NLP techniques, so it is reasonable to assume that due to its reliance on semantic information contained in the dataset significant benefits could be gained from using TLMs and CWRs. However, building a model which uses these more advanced techniques to provide improved performance and more refined insights would be non-trivial. While gains could be expected from fine-tuning some of the feature extraction techniques on labeled datasets, the value added may not be significant enough to impact the results. |
| **Dong et al. 2018** | This study utilizes social media data for fraud detection (as opposed to financial statements or financial data). The authors propose a framework, grounded in systemic functional linguistics theory, and a matched dataset of 64 fraudulent and 64 nonfraudulent firms to predict fraud. The framework extracts text-based features via sentiment analysis, emotion analysis, lexical features and topic modeling for input into an SVM classifier. The sentiment analysis is performed using a sentiment dictionary for financial applications, the emotions are taken from the LIWC dictionary, the lexical features are extracted via TF-IDF and LDA is used for feature extraction via topic modeling. Network-based features are also extracted. | This is another text-based model that uses a multi-method framework to build a predictive model. Because the model is intended to predict fraud detection directly, it could be used to train directly on the documents in the corpus to predict fraud. However, this is unlikely to be effective because the strength of indicators of fraud in social media data is likely to vary substantially. It is possible some novel technique could be used to leverage TLM fine-tuning directly, although a possible technique for this is not apparent to us. However, we feel that it is unlikely that CWRs could not add value due to their superior ability to capture latent semantic features of language, but this still may not significantly affect the results. |
| **Ghiassi et al. 2016** | This study focuses on the development of a novel sentiment analysis method for brand related Twitter data. The technique involves targeted sentiment analysis using supervised feature engineering combined with a dynamic neural network architecture. The technique is evaluated for both three-class and five-class sentiment classification and outperforms two state-of-the-art (SOTA) Twitter sentiment analysis systems. | We expect that fine-tuning using the two brand-related Twitter datasets would be able to outperform this technique. Because this is a strong model, the size of the TLM and the pretraining data would likely make a difference in whether or not a model could outperform the model from this study. For example, RoBERTa and T5 11B may be able to outperform this model with generic pretrained data, while BERT may not be able to outperform this model or may only do so when pretrained on Twitter data. It is possible our suspicions are incorrect, but we are confident that performance could be improved using a complex ensemble technique like Naseem et al. [2020]. |
| **Huang et al. 2017** | This study explores the impact of online comments from a variety of sources and platforms on a film's box office performance over time. The authors used text mining/sentiment analysis on over 1,500 expert and consumer reviews for films released over an entire year. A naive Bayes sentiment model was trained by vectorizing documents and using assigned numerical ratings as the target (the training data was randomly sampled from the parent dataset). | This may be one of the studies which stands to benefit most from using either TLMs or CWRs. It is such a good candidate because it has a unique topic-specific dataset which can be used for training, and it uses a very traditional technique for training the model that was originally used. A language model could be operationalized by either training TLMs directly for the task of sentiment analysis or by using TLMs and CWRs to create contextual document vectors that can then be used as features for training a sentiment model, e.g., [Adhikari]. Because sentiment is so critical to this model, the results stand to be impacted significantly. |
| **Kumar et al. 2019** | This study focuses on the detection of fake online reviews due to the significant role they play in consumers' online purchasing decisions. The authors propose an entirely unsupervised, novel, hierarchical approach that involves deriving key feature distributions characterizing reviewers' behavior and using the combined distributions in a finite mixture model for anomaly detection (i.e. fake reviewers). Only frequency-based text mining is used in this study (i.e., word count). | Certainly, TLMs could be fine-tuned to recognize fake reviews, although the quality of the results may not be strong. However, it is likely that generative language models will be used in the future to create fake reviews, and TLMs should be able to perform well at detecting text generated from other TLMs. It is possible that CWRs or TLMs could be used to generate novel frameworks which could outperform this model, but how to do this is not clear or straightforward to us. This is an interesting topic for future research, but it is likely that this is one of the few studies in this review that would not benefit even marginally from the new techniques. |
| **Li et al. 2016** | This study presents the AZSecure text mining system for identifying and profiling key underground economy sellers of credit/debit cards in order to protect consumers from cybercrime. The system uses machine translation for preprocessing foreign language data and uses both sentiment analysis and topic modeling (LDA) for feature extraction. The resulting features are used for training a max entropy classifier. | Because we do not have good knowledge of the data used we are unable to say whether or not we feel that a TLM could be fine-tuned to directly classify either malware or stolen data advertisements, although it would be possible if there were strong signals of criminality in each advertisement. Regardless, this, as with most models in this literature review, could utilize the strong semantic features that are captured with CWRs and leverage them to improve performance. Also, due to the multilingual nature of the data, we are confident that errors exist from the machine translation and that this element could be improved with multilingual TLMs. |
| Mai et al. 2018 | This study explores the determinants of Bitcoin's valuation by exploring the dynamic interactions between social media and valuation using text mining and vector error correction models. Social media data collected from a Bitcoin discussion board and Twitter was analyzed using sentiment analysis (using a financial sentiment dictionary; news data was also included) and coupled with Bitcoin pricing data. Features were used that were computed as the number of positive and negative posts/Tweets. These were combined with other features like market value, commodity value, investor sentiment, news sentiment, etc. for input into a model. | This study stands to potentially gain moderately to significantly because sentiment analysis plays such a big role in the study, and the research deals with text data from two unique contexts. It is unlikely that the financial sentiment dictionary used was optimal for either the Bitcoin community posts or the Bitcoin hashtag Tweets. Due to the unique contexts, models trained on crowd labeled data from these data sources could likely improve performance which would have a direct impact on the results of the study. Large language models like GPT-3 have shown strong performance on Twitter sentiment analysis. Moreover, it would be interesting to extend this study to assess other cultures' sentiment due to the Global nature of Bitcoin using multilingual TLMs. |
| Saifee et al. 2019 | This study was intended to explore questions regarding the relationships between adherence to clinical guidelines and the use of electronic health records, and physicians' online reputations. Data was obtained from multiple sources and the main dependent variables were constructed from the physicians' online reviews: sentiment scores and overall rating (from the reviews; seven different dimensions). Sentiment analysis was conducted using the SentimentR package in R which uses an unspecified generic sentiment analysis method that uses a context window for identifying valence shifters. | Because sentiment is one of the primary dependent variables in this study, it is likely that the results could be impacted by improving the quality of sentiment analysis. While the method used is robust for a wide variety of tasks, it is not specialized for the specific task. Using crowd sourcing to label the sentiment of different posts (perhaps n = 10,000) for training a sentiment model, likely as part of a multitask TLM (including multiple sentiment tasks), could be expected to improve performance of the sentiment analysis module. Whether this would have a significant effect on the results and the quality of the insights gained from the study is unclear, however, and it is possible that there may be no broader gains. |
| Samtani et al. 2017 | This study examines a novel framework for cyber threat intelligence intended to provide proactive advice for better understanding the present threats in hacker communities. The authors utilize a multimethod approach that leverages a variety of techniques to incorporate a range of relevant data, including posts from hacker community forums, as part of the complex cyber threat intelligence system. LDA is applied to attachment and tutorial posts for feature extraction. | This study is very interesting when considering future directions for TLM and CWRs in IS research. For example, the hacker community is a global community, which includes large communities using different languages. Thus, cross-cultural exploration of this community through multilingual TLMs could improve work like this and lead to better insights. Separately, TLMs have been demonstrated to be effective at generating or correcting source code for some programming languages. However, source code generation isn't necessary for applications like this where latent feature extraction from source code can be used as valuable features for a model. |
| Shi, et al. 2017 | This study involves a novel data mining approach to risk factor identification and analysis in safety management systems using data from the Aviation Safety Reporting System (i.e., FAA). The new technique is intended to overcome challenges of missing data and labor-intensive manual analysis of risk factors. LSA/SVD was used for topic modeling/feature extraction to create features to use as inputs for training classifiers (naive Bayes, decision trees and boosting). | This study is interesting because it offers a good opportunity to build an end-to-end classifier by only fine-tuning a TLM. Because incidents of different classes are likely to have significant semantic differences, it is likely that a TLM developed from simple fine-tuning on labeled incidents would be able to offer good performance. We suspect that with a large enough TLM (e.g., T5 or GPT-3) performance would be significantly better than this study, and all without significant effort in model development. So, this is an example that stands to benefit greatly from the new techniques. We further suspect few-shot performance from GPT-3 or other large language models would be beneficial for this task and may lead to state-of-the-art results with minimal effort. |
| Siering et al. 2016 | This study demonstrates a novel technique for detecting deception and fraud in crowdsourcing platforms by analyzing a sample of fraudulent and nonfraudulent proposals posted on a leading crowdfunding platform. The multimethod framework involves assessing 1) content-based cues via a bag-of-words approach by creating document vectors through TF-IDF and 2) linguistic cues through category-based measures by using the General Inquirer dictionary. Numerous machine learning classifiers are used including SVMs, naive Bayes, k-nearest neighbors, decision trees and all of these together in ensembles. | This is another case that is difficult to project the true value of the new techniques because the semantic content of the dataset is not clear. If there are clear and strong signals of deception and fraud then an end-to-end classifier could be trained, although this is unlikely. What is more likely is that complex systems involving TLMs or CWRs can be designed that are able to improve upon the performance of the model from this study. Regardless, because this is a predictive model the performance is important and some combination of TLMs or CWRs will be able to significantly increase predictive ability on this task, albeit doing so may be nontrivial and may require substantial expertise and experience with NLP and TLMs. |
| **Van Osch et al. 2018** | This study explores the ways in which enterprise social media visibility is used strategically and its implications for cross-boundary knowledge sharing by examining three types of boundary-spanning interactions. A machine learning model was used to generate measures for each of these three dependent variables using log and content data from enterprise social media-based workgroups. The first step involved labeling text data for training an SVM for text classification. Network features were used as independent variables for the statistical model. | For text classification, it is likely that performance gains could be realized because the documents in question are of relatively short token lengths. It is uncertain whether basic models such as BERT could be trained to improve upon the results from this study, but it is highly likely that by simply fine-tuning on the labeled dataset used in the study, likely in a multitask framework, performance could be improved with models designed for deep transfer learning such as T5. CWRs could likely also be a solid avenue for improving classification. Because the text classification was used for the dependent variables, the impact on the results could potentially be significant. |
| **Velichety et al. 2019** | This study involves the development of a process for identifying features for quality assessment in large dataset and a method to assess peer-produced content quality in knowledge repositories, i.e., Wikipedia, using a design science framework. Sentiment analysis (using a dictionary) and topic modeling (using LDA) were used to extract text-based features and were combined with a variety of metadata features specific to the dataset and network features that were used as inputs to the statistical model. The paper also involved Hadoop because it used such a large amount of data. | Despite this study using so many different elements in the novel framework, new techniques such as TLMs or CWRs still stand to offer substantive improvements due to the importance of the features for which sentiment analysis and topic modeling are used to create. Because the dataset is general and not domain specific, pretrained TLMs, such as XLNet [Yang 2019], already achieve SOTA performance on sentiment analysis. It may not produce results of the same fidelity, but it would be interesting to fine-tune a large TLM directly on Wikipedia content to predict the quality of articles. By using this prediction with other features, SOTA performance is likely. |
| **Wang et al. 2020** | This work involves a novel text mining technique for end-to-end extraction of semantic features for soft information contained in descriptive loan texts submitted by borrowers. The technique creates representations in a latent feature space and clusters related terms, labeling the clusters based on their semantic soft factor characteristics. The clusters, called semantic cliques, are created using a novel algorithm. Evaluation shows that the soft factor clusters act as features in a predictive model to significantly improve credit risk evaluation. | Because this uses word representations directly, without having to generate document vectors/representations, it is a great candidate for using CWRs. Simply using these alternative representations in the current model would almost certainly improve predictive performance due to the valuable semantics of context that are encoded in CWRs. Similar to other studies, it would be interesting to fine-tune a large TLM directly on the descriptive loan texts to predict the risk of default. This problem differs from some other problems discussed in this review, and we feel it has greater potential via this approach. |
| **Yoo et al. 2019** | This study evaluates the propagation of user-generated content in social networks, and how diffusion for content is impacted by the diffusion of similar content. The analysis involves a notion of information cascades and parallel cascades that were first used to investigate rumor propagation in Twitter via retweets. To these ends, the authors use Twitter data concerning natural disasters requiring aid from humanitarian organizations. Text mining via the SimHash algorithm was used to determine similarity between Tweets in parallel cascades that contained similar information. | We are not knowledgeable about the use of cascades for analysis of the diffusion of user content in Twitter data, but BERT has been used to create document vectors which have outperformed SimHash for tasks related to the deduplication of scholarly documents [Gyawali 2020]. Thus, we are confident that TLMs could be used to generate CWRs for aggregation into document representations that could outperform the similarity measurement technique applied here. Because Twitter data is involved, a TLM pretrained on Twitter data may perform best. Due to the role of parallel cascade in this research, the impact could be substantive. |
| **Zhang et al. 2016** | This study also concerns the issue of fake online reviews, but it is intended to compare the significance of nonverbal behavioral features and verbal behavioral features in predictive models for identifying fake reviews. The study uses text mining for extracting all frequency-based features from documents to use as inputs for an SVM classifier. They find that the inclusion of nonverbal features into models can improve performance of fake review detection systems. | Because the text mining techniques used here are all for feature extraction and are all frequency based, it is likely that the rich semantic information captured in CWRs could be leveraged to substantially improve fake review detection using all verbal features, though combined methods would probably still perform best. It would also be interesting to train the model directly to predict fake reviews, similar to other cases. While it is likely combined methods will be optimal for the near-term, it is possible that continued scaling of TLMs could lead to superior performance at some point. |
| **Zhou et al. 2018** | This study considers the impact of product reviews from the perspective of product developers through a big data approach that analyzes product review notes and a mobile app dataset comprised of over three million reviews. The authors use TF-IDF in an SVD-based semantic similarity method to quantify customer agility. This similarity measure marginally outperforms Word2Vec (GloVe, which would seem the more appropriate word representation comparison given the use of SVDs, was not shown). | Because the technique only marginally outperforms more rudiment word representations, we are very confident that CWRs could be used to improve performance of this model, and, consequently, have significant implications for the results and insights gleaned from this study. Using CWRs in this way is one manner that can be used to evaluate the research questions of concern here, but alternate ways may also be possible with TLM fine-tuning or different uses of CWRs. |

## Table A3: MIS Quarterly Survey Summaries and Recommendations

| Study | Summary | Recommendations (*e.g.*, future work) |
|---|---|---|
| Abbasi et al. 2018 | Using design science, this study proposes a language-action perspective (LAP) text analytics framework to support sense-making in online discourse. The proposed framework has three components: 1. conversation disentanglement 2. coherence analysis and 3. message speech act classification. The system uses numerous forms of classification to extract different system/linguistic/conversational structure features in order to feed into a "speech act theory (SAT)" tree for classification. | The proposed model is heavily grounded in theory, and consequently has a large number of components. Thus, it is hard to describe specifically how TLMs or CWRs could be used to improve this model, and it is also difficult to determine how much improvement the framework would see if using some of these advanced techniques. However, we feel that CWRs could certainly be used for some portions of this model, *e.g.*, semantic features could be extracted rather than just linguistic features. This example might not make a significant difference, but we feel that this could at least modestly improve the model. |
| Arazy et al. 2020 | This study concerns peer-produced digital artifacts from online communities and analyzes the evolution of wiki articles to assess the evolution of the artifacts over time. The study uses the ORES algorithm for extracting a feature estimating the quality of each wiki article. This and other features were used in regression models. | This article can clearly be improved by TLMs and CWRs as the authors explicitly suggest that the feature space could be represented more sophisticatedly using text analytics techniques and more advanced knowledge representations. They further suggest that the patterns observed in this study are a lower bound. Thus, we recommend using the more powerful methods that are the focus of this study to improve this framework in future work. |
| Bapna et al. 2019 | This study explores how firms can nurture online brand communities in social media that can bring value to firms. The authors used manual coders to extract features from posts via Mturk and also used LDA to identify the types of posts that were used by firms which might not be associated with engagement. The extracted features were used as input for the analytical model. | The coding done in this study by Mturk workers is superior to anything that TLMs or CWRs can currently provide. However, using crowd workers does not scale to big data and is only effective on modestly sized datasets unless the research team is very well funded. Thus, we do not expect significant improvements to the results of this study to be likely, although marginal to modest improvements may be realized from using CWRs instead of LDA. |
| Benjamin et al. 2019 | This study offers a framework for guiding scholars in Darknet research so that they may contribute to the body of literature: Darknet Identification, Collection, Evaluation, with-Ethics (DICE-E). Data found in the darknet is largely comparable to that found in more traditional forums such as social media and includes a lot of text content. However, while they suggest the use of a range of text mining techniques for content analysis, their example only uses parsing and no other text mining techniques. | The text mining techniques used in this are not very complex, but TLMs and CWRs do stand to benefit this research significantly. Because the Darknet is a global network and features actors from numerous countries, it seems logical that multilingual TLMs would be useful. In general, all of the methods suggested for content analysis could likely be improved with the new techniques, and especially so with multilingual models. This presents a good opportunity for future work for those interested in exploring Darknet research. |
| Chau et al. 2020 | This study focuses on identifying emotionally distressed people through their social media posts. Design science is used to develop a new system for identifying and classifying content from distressed individuals. Due to the nature of the relevant content, the approach heavily uses sentiment and affect analysis (lexicon-based). Two classifiers are combined in an ensemble - one an ML classifier and the other a rule-based classifier. The ML classifier utilizes feature extraction and an SVM classifier. | Because this is primarily a text analytics model utilizing more traditional techniques, it has substantial potential for improvement by adopting TLMs and CWRs. As it relies heavily on sentiment analysis and affect analysis, there are several options that could be used for improving performance. One would be to simply use a pretrained and already fine-tuned sentiment TLM that already achieves SOTA performance such as those available in Hugging Face's transformers library [Wolf 2019]. Alternately, due to the nature of this problem and the impact of accuracy, a crowd labeled emotion dataset could be trained via a large TLM. |
| Deng et al. 2018 | This study shows that the effect of microblog sentiment on stock returns is both statistically and economically significant at the hour level. It uses sentiment analysis and an autoregressive model for a very large dataset of StockTwits messages (~18M tweets identified by a cashtag, *i.e.*, "$") that spanned four years. The sentiment analysis used is SentiStrength and a vector autoregression model was used for the time series analysis. | This study is focused on sentiment analysis as it explores the impact of sentiment on stock trading. Due to this, similar to the Chau et al. 2020, the work has substantial potential to see improvement from the adoption of TLMs and CWRs. Again, similar to the previous study, already fine-tuned sentiment models could be used, or, even better, a crowd labeled StockTwits dataset could be created and used to fine-tune a large TLM. We feel that the latter would greatly benefit the results and is a good direction for future research. |
| Gong et al. 2017 | This study proposes an automatic way of examining keyword ambiguity based on probabilistic topic models in order to explore its effect on search advertising performance. The authors utilize a hierarchical Bayesian approach considering topic-specific effects and nonlinear position effects for modeling click-through rate and ad position (rank). Topic modeling via LDA is used to generate topics for feature extraction. | Since LDA is simply used for feature extraction we feel that it is likely CWRs could be used in a straightforward manner to generate latent features which could be used effectively in the model to at the least modestly improve performance. Because search queries are very short and contain very strong semantic signals, it is likely that TLMs could be used with fine-tuning to create much stronger means of conducting this type of classification. |
| Huang, K.Y., et al. 2019 | The paper examines the relationships between dimensions of social capital and the provision of support and companionship in healthcare virtual support communities. A model is developed and tested on data from three virtual health support communities. Messages in the communities were manually coded and separate text classifiers for each group to identify whether messages were for support or companionship and further whether they were informational or emotional support. A model was developed based on the results of the classifiers in combination with other relevant independent variables. | This is an excellent opportunity for using TLMs directly with fine-tuning. It is ideal because the training dataset sizes are small and because the quality of the classification is critical to the results of the study. BERT would be straightforward, and could likely be used to improve this for free in a Google Colab demonstration notebook. This is also a good example for using T5 in a multitask transfer learning capacity because T5 could be trained simultaneously on the various different classification tasks involved in this study by simply appending a task label to be beginning of each input. |
| Huang, N. et al. 2016 | This study examines two natural experiments at leading online review platforms and data from Facebook to assess the impact of social network integration. The authors use Linguistic Inquiry and Word Count (LIWC) to construct frequency-based measures of linguistic content for sentiment analysis and feature extraction. A panel dataset of matched restaurants is created using these features in a difference-in-differences model assessing the impact of social network integration. | While frequently used in psychology literature and in previous IS research, LIWC is less than optimal given recent advances and this model could be improved by using TLMs in a custom implementation specific to the topic. However, we feel that generalizable affective models specific to IS or psychology research are desirable due to the popularity of LIWC and other models common in the IS literature. We also recognize that it may be non-trivial to develop a model that can be used across subdomains in these areas to outperform existing methods. Thus, we feel it in the interest of all IS researchers for future work to focus on developing TLMs specific to the IS and psychology research domains. |
| Li, J. et al. 2020 | This study uses a design science approach and proposes a behavioral ontology learning from text (BOLT) design framework for supporting researchers during the processing of behavioral knowledge. From this an instantiation - a search engine, TheoryOn - is developed that allows researchers to search directly for constructs, construct relationships, antecedents and consequents. The framework and search engine were rigorously evaluated through a series of data mining experiments. It uses a broad range of techniques, including entity extraction, text classification, word representations and LSTMs. | This is one of the most advanced models, yet, as the authors state, "according to the BOLT framework, the performance of theoretical relationship extraction was heavily dependent on the accuracy of variable extraction," so we still feel TLMs and CWRs stand to benefit TheoryOn substantially. With respect to TheoryOn, an easy but likely substantive improvement could be realized by simply utilizing CWRs as opposed to word2vec representations because the contextual semantic information captured in these representations is richer than earlier neural representations. |
| Liu, X. et al. 2019 | This study proposes using an interdisciplinary lens to combine deep learning with themes emphasized in IS and healthcare informatics to examine user engagement with encoded medical info from YouTube videos. The authors use a Bi-LSTM to identify medical terms and classify videos based on how much medical information is encoded. They then use PCA on the video data to analyze dimensions of user engagement. Robustness checks were conducted with caption mining and image mining. | LSTMs have had much success at tasks related to image and video captioning, but it appears that TLMs and CWRs are now being used effectively at these tasks also. Work from Qi et al. [2020] has demonstrated success with ImageBERT on related tasks, and Cho et al. [2020] have demonstrated the use of X-LXMERT for image captioning and visual Q&A. So, it seems that presently classification of videos may not be able to be improved beyond SOTA using TLMs, but it is likely that soon there will be TLM-based systems which are capable of this as well. |
| Rhue & Sundararajan 2019 | This study examines how the dynamics of social commerce affect visibility choices of consumers' purchasing behaviors by analyzing data from an online "Twitter for credit card purchases." The authors use fixed effects and matching for the statistical model and they use a dictionary-based sentiment analysis for extracting a feature to use as input for this model. | Since sentiment analysis is the only text mining technique used in this study, and because it was only used as a single feature in the model, we do feel that TLMs could bring modest improvement to this study, but we do not feel that significant gains, as for other studies, are possible. For sentiment analysis either pretrained models or domain-specific fine-tuned sentiment models could be used. Because this study deals with reviews for a range of products it is likely that the performance of pretrained models like XLNet would be suitable for improving overall model performance, particularly because a dictionary-based technique was used originally. |
| Shi et al. 2016 | This study involves the collection of a dataset from CrunchBase to evaluate proposed analytic approach for measuring firms' dyadic business proximity. The authors use LDA to represent a firm's textual description as a probability distribution over a set of underlying topic which are interpreted as aspects of the business. The "business proximity" is simply a measure of the distance between two firms topic distributions. | This is a case that can likely be improved with CWRs as the novel model here attempts to use distributional semantics to measure business proximity. CWRs offer much richer probabilistic semantic representations of words, and thus, could likely be used in a manner to generate richer and more accurate representations of the aspects of firms' business. However, while we feel confident that gains could be realized in this manner it may not be straightforward due to the need for a novel measure that this work demonstrates, but we do feel that work using CWRs in this way could be valuable. |
| Shin et al. 2019 | This study proposes several visual and textual social media features that can be used for analysis of pervasiveness of content. The authors then deploy deep learning and text mining to operationalize the new features in a scalable and systematic manner. The features are validated via the Mturk and two case studies from Tumblr. They use topic modeling, word embeddings and feature extraction for text in combination with convolutional neural networks for images. | Due to the heavy reliance on representations from word2vec, the use of CWRs can be expected to substantially improve performance of this model. While this study does not utilized these more advanced representations, the authors do cite BERT as an example of the value of using rich word embeddings for cases of sparse data. We interpret this as a signal that soon we will begin seeing these methods utilized in IS research, even if they are only used by a minority in the beginning. Further. because this method uses both visual and text analytics it is likely that some of the multi-modal models like ImageBERT [Qi 2020] or X-LXMERT [Cho 2020] could also be beneficial here. |
| Wu et al. 2019 | This study demonstrates a new avenue of research on how social theories can be operationalized in e-commerce operations by building on regulatory focus theory (RFT). The authors center on customer regulatory focus and employ a sentiment analysis technique dubbed "RF discovery" via a custom adaptation of the SentiStrength platform/application. Sentiment analysis is used for multiple elements of this paper and is the primary feature used in analysis. | Because this study relies so heavily on sentiment analysis, it could benefit from the use of TLMs. The most obvious option would be to use pretrained models like XLNet that already perform well on general sentiment analysis tasks. A pretrained model could also be fine-tuned on a crowd-labeled dataset specific to e-commerce. Or, a large e-commerce dataset could be used for unsupervised pretraining of a domain-specific BERT (*e.g.*, "e-CBERT") which could then be further fine-tuned specifically for sentiment analysis on different categories of reviews. An interesting alternative would be to train a domain-specific T5 which could then be used for multitask training of sentiment analysis for different product categories. |
| Yue et al. 2019 | This study looks at online hacker forums to analyze the impact of information shared in online discussions on the frequency of DDOS attacks. A field dataset collected from several sources is analyzed using a panel fixed effects model and data regarding the DDOS attacks. Content analysis is conducted using two text mining techniques: LDA for topic modeling and tf-idf supervised classification. The authors found that the content analysis didn't have a sizable effect on the model for English language posts and chose not to expend the necessary resources to analyze content for Russian and Chinese language posts for the majority of the study. | This study is interesting for two reasons. First, the text mining techniques used did not provide a strong enough effect to include in the statistical model. This is interesting because of the significance of errors and uncertainty in methods used for text mining feature generation [Yang 2018]. Thus, we feel that these newer methods may certainly impact the value of these features extracted from the posts and we think it would be very informative to learn of whether or not stronger methods impacted the results of the study. Second, this study involves multiple languages, and thus, the costs associated with analysis of the data in other languages could be reduced with multilingual TLMs, and this is a good case for exploring their use. |
| Zhang et al 2016 | This paper proposes an audience selection framework for analyzing brand-brand networks for online brand advertising using user activity on social media. The authors extract and analyze brand-brand networks using network analytics to propose a framework combing text and network analytics for finding target audiences - the network analytics techniques are combined with sentiment analysis using a technique previously demonstrated to perform well for social media content. MapReduce is used to process big data for this study. | This study focuses more on network analytics that text mining, so the impact of TLMs or CWRs on the results is likely to be marginal to modest at best. While this may be the case, if one were to use TLMs they could simply use XLNet or another strong pretrained TLM which performs well for sentiment. It is likely not worth the effort to develop more sophisticated sentiment analysis techniques using TLMs for cases such as this. |
| Zhang & Ram 2020. | The study introduces a data-driven framework that integrates a variety of machine learning techniques and performs an empirical analysis in order to aid in developing guidance for asthma management plans and interventions for specific subpopulations. A previous method from the authors is used for feature extraction and text classification to identify potential asthma patients. Sequential pattern mining is first used for extracting features which are ultimately used as inputs for the analytical model. | The use of text mining in this study is not for a critical part of the study, but only for classifying potential asthma patients. Consequently, alternative methods with modest improvements for the task are unlikely to have a major impact on the results, but TLMs and CWRs could be used for this purpose. It is probable that the most straightforward way to implement these techniques would be by directly fine-tuning a TLM for this classification task using labeled training data. |

# REFERENCES

Abbasi, A., Zhou, Y., Deng, S., and Zhang, P., 2018. "Text analytics to support sense-making in social media: A language-action perspective," *MIS Quarterly*, *42*(2), pp. 427-464.

Abbasi, A., Li, J., Adjeroh, D., Abate, M., and Zheng, W. 2019. "Don't Mention It? Analyzing User-Generated Content Signals for Early Adverse Event Warnings," *Information Systems Research*, *30*(3), pp.1007-1028.

Adamopoulos, P., Ghose, A., and Todri, V. 2018. "The impact of user personality traits on word of mouth: Text-mining social media platforms," *Information Systems Research*, *29*(3), pp.612-640.

Adhikari, A., Ram, A., Tang, R., and Lin, J. 2019. "Docbert: Bert for document classification," (*arXiv preprint arXiv:1904.08398*).

Arazy, O., Lindberg, A., Rezaei, M., and Samorani, M. 2020. "The Evolutionary Trajectories of Peer-Produced Artifacts: Group Composition, the Trajectories' Exploration, and the Quality of Artifacts," *MIS Quarterly*, *44*.

Bapna, S., Benner, M.J., and Qiu, L. 2019. "Nurturing Online Communities: An Empirical Investigation," *MIS Quarterly*, *43*(2), pp. 425-452.

Benjamin, V., Valacich, J.S., and Chen, H. 2019. "DICE-E: A Framework for Conducting Darknet Identification, Collection, Evaluation with Ethics," *MIS Quarterly*, *43*(1).

Blohm, I., Riedl, C., Füller, J., and Leimeister, J.M. 2016. "Rate or trade? Identifying winning ideas in open idea sourcing," *Information Systems Research*, *27*(1), pp.27-48.

Chau, M., Li, T.M., Wong, P.W., Xu, J.J., Yip, P.S. and Chen, H. 2020. "Finding People with Emotional Distress in Online Social Media: A Design Combining Machine Learning and Rule-Based Classification," *MIS Quarterly*, *44*(2), pp. 933-953.

Chen, K., Li, X., Luo, P., and Zhao, J.L. 2020. "News-Induced Dynamic Networks for Market Signaling: Understanding Impact of News on Firm Equity Value," *Information Systems Research*.

Chen, L., Baird, A., and Straub, D. 2019. "Fostering Participant Health Knowledge and Attitudes: An Econometric Study of a Chronic Disease-Focused Online Health Community," *Journal of Management Information Systems*, *36*(1), pp.194-229.

Chen, W., Gu, B., Ye, Q., and Zhu, K. 2019. "Measuring and managing the externality of managerial responses to online customer reviews," *Information Systems Research*, *30*(1), pp. 81-96.

Cho, H., Knijnenburg, B., Kobsa, A., and Li, Y. 2018. "Collective Privacy Management in Social Media: A Cross-Cultural Validation," *ACM Transactions on Computer-Human Interaction* 25(3), pp. 1-33.

Chung, S., Animesh, A., Han, K., and Pinsonneault, A. 2020. "Financial returns to firms' communication actions on firm-initiated social media: evidence from Facebook business pages," *Information Systems Research*, *31*(1), pp.258-285.

Deng, S., Huang, Z., Sinha, A., and Zhao, H. 2018. "The Interaction Between Microblog Sentiment and Stock Return: An Empirical Examination," *MIS Quarterly*, *42*(3), pp. 895-918.

Dong, W., Liao, S., and Zhang, Z. 2018. "Leveraging Financial Social Media Data for Corporate Fraud Detection," *Journal of Management Information Systems*, *35*(2), pp.461-487.

Ghiassi, M., Zimbra, D., and Lee, S., 2016. "Targeted Twitter Sentiment Analysis for Brands Using Supervised Feature Engineering and the Dynamic Architecture for Artificial Neural Networks," *Journal of Management Information Systems*, *33*(4), pp.1034-1058.

Gong, J., Abhishek, V. and Li, B., 2017. Examining the Impact of Keyword Ambiguity on Search Advertising Performance: A Topic Model Approach," *MIS Quarterly* *43*(3), pp. 805-829.

Gyawali, B., Anastasiou, L., and Knoth, P. 2020. "Deduplication of Scholarly Documents using Locality Sensitive Hashing and Word Embeddings," *Proceedings of 12th Language Resources and Evaluation Conference*, pp. 901-910.

Heimbach, I., and Hinz, O. 2018. "The impact of sharing mechanism design on content sharing in online social networks," *Information Systems Research*, 29(3), pp. 592-611.

Huang, J., Boh, W., and Goh, K. 2017. "A temporal study of the effects of online opinions: Inform-ation sources matter," *Journal of Management Information Systems*, 34(4), pp. 1169-1202.

Huang, N., Hong, Y., and Burtch, G. 2017. "Social network integration and user content generation: Evidence from natural experiments," *MIS Quarterly*, 41(4), pp. 1035-1058.

Hwang, E. H., Singh, P. V., and Argote, L. 2019. "Jack of all, master of some: Information network and innovation in crowdsourcing communities," *Information Systems Research*, 30(2) pp. 389-410.

Khern-am-nuai, W, Kannan, K., and Ghasemkhani, H. 2018. "Extrinsic Versus Intrinsic Rewards for Contributing Reviews in an Online Platform," *Information Systems Research*, 29(4), pp. 871-892.

Kumar, N., Venugopal, D., Qiu, L., and Kumar, S. 2019. "Detecting Anomalous Online Reviewers: an Unsupervised Approach Using Mixture Models," *Journal of Management Information Systems*, 36(4), pp. 1313-1346.

Kraus, M., and Feuerriegel, S. 2017. "Decision support from financial disclosures with deep neural networks and transfer learning," *Decision Support Systems*, 104, pp. 38-48.

Lappas, T., Sabnis, G., and Valkanas, G. 2016. "The Impact of Fake Reviews on Online Visibility: A Vulnerability Assessment of the Hotel Industry," *Information Systems Research*, 27(4), pp. 940-961.

Lee, G.M., He, S., Lee, J., and Whinston, A.B. 2020. "Matching Mobile Applications for Cross-Promotion." *Information Systems Research*, *31*(3), pp.865-891.

Li, J., Larsen, K., and Abbasi, A. 2020. "TheoryOn: A Design Framework and System for Unlocking Behavioral Knowledge Through Ontology Learning," *MIS Quarterly*, 44(4), pp. 1733-1772.

Li, W., Chen, H., and Nunamaker, J. F. 2016. "Identifying and Profiling Key Sellers in Cyber Carding Community: AZSecure Text Mining System," *Journal of Management Information Systems*, 33(4), pp. 1059-1086.

Liu, X., Zhang, B., Susarla, A., and Padman, R. 2020a. "Go to YouTube and Call Me in the Morning: Use of Social Media for Chronic Conditions," *MIS Quarterly*, 44(1b), pp. 257-283.

Liu, X., Wang, G. A., Fan, W., and Zhang, Z. 2020b. "Finding Useful Solutions in Online Knowledge Communities: A Theory-Driven Design and Multilevel Analysis," *Information Systems Research*, 31(3).

Liu, Y., Pant, G., and Sheng, O. R. L. 2020c. "Predicting Labor Market Competition: Leveraging Interfirm Network and Employee Skills," *Information Systems Research, forthcoming*.

Mai, F., Shan, Z., Bai, Q., Wang, X., and Chiang, R. H. L. 2018. "How does social media impact Bitcoin value? A Test of the Silent Majority Hypothesis," *Journal of Management Information Systems*, 35(1), pp. 19-52.

Mejia, J., Mankad, S., and Gopal, A. 2019. "A for Effort? Using the Crowd to Identify Moral Hazard in New York City Restaurant Hygiene Inspections," *Information Systems Research*, 30(4), pp. 1363-1386.

Mousavi, R., and Gu, B. 2019. "The Impact of Twitter Adoption on Lawmakers' Voting Orientations," *Information Systems Research*, 30(1), pp. 133-153.

Mousavi, R., Johar, M., and Mookerjee, V. J. "The Voice of the Customer: Managing Customer Care in Twitter," *Information Systems Research*, 31(2), pp. 340-360.

Naseem, U., Razzak, I., Musial, K. and Imran, M. 2020. "Transformer based deep intelligent contextual embedding for twitter sentiment analysis," *Future Generation Computer Systems*, 113, pp. 58-69.

Pan, Y., Huang, P., and Gopal, A. 2019. "Storm Clouds on the Horizon? New Entry Threats and R&D Investments in the US IT Industry," *Information Systems Research*, 30(2), pp. 540-562.

Qi, D., Su, L., Song, J., Cui, E., Bharti, T., and Sacheti, A., 2020. "Imagebert: Cross-Modal Pre-Training with Large-Scale Weak-Supervised Image-Text Data," (*arXiv preprint arXiv:2001.07966*).

Rhue, L., and Sundararajan, A. 2019. "Playing to the Crowd? Digital Visibility and the Social Dynamics of Purchase Disclosure," *MIS Quarterly*, 43(4), pp. 1127-1141.

Saifee, D. H., Bardhan, I.R., Lahiri, A., and Zheng, Z. 2019. "Adherence to Clinical Guidelines, Electronic Health Record Use, and Online Reviews," *Journal of Management Information Systems* 36(4), pp. 1071-1104.

Samtani, S., Chinn, R, Chen, H., and Nunamaker, J. F. 2017. "Exploring emerging hacker assets and key hackers for proactive cyber threat intelligence," *Journal of Management Information Systems*, 34(4), pp. 1023-1053.

Shi, D., Guan, J., Zurada, J., and Manikas, A. 2017. "A Data-Mining Approach to Identification of Risk Factors in Safety Management Systems," *Journal of Management Information Systems*, 34(4), pp. 1054-1081.

Shi, Z., Lee, G. M., and Whinston, A. B. 2016. "Toward a Better Measure of Business Proximity: Topic Modeling for Industry Intelligence," *MIS Quarterly*, 40(4), pp. 1035-1056.

Shin, D., He, S., Lee, G.M., Whinston, A.B., Cetintas, S. and Lee, K.C. 2019. "Enhancing Social Media Analysis with Visual Data Analytics: A Deep Learning Approach," *MIS Quarterly*, 2020.

Siering, M., Koch, J. and Deokar, A. 2016. "Detecting Fraudulent Behavior on Crowdfunding Platforms: The Role of Linguistic and Content-Based Cues in Static and Dynamic Contexts," *Journal of Management Information Systems*, 33(2), pp. 421-455.

Van Osch, W., and Steinfield, C. W. 2018. "Strategic Visibility in Enterprise Social Media: Implications for Network Formation and Boundary Spanning," *Journal of Management Information Systems*, 35(2), pp. 647-682.

Velichety, S., Ram, S., and Bockstedt, J. "Quality Assessment of Peer-Produced Content in Knowledge Repositories Using Development and Coordination Activities," *Journal of Management Information Systems*, 36(2), pp. 478-512.

Wang, A., Singh, A., Michael, J., Hill, F., Levy, O., and Bowman, S.R. 2018a. "GLUE: A Multi-Task Benchmark and Analysis Platform for Natural Language Understanding," *EMNLP Workshop on BlackBox NLP*, pp 353-355.

Wang, Q., Li, B., and Singh, P. V. 2018b. "Copycats vs. Original Mobile Apps: A Machine Learning Copycat-Detection Method and Empirical Analysis," *Information Systems Research*, 29(2), pp. 273-291.

Wang, Z., Jiang, C., Zhao, H., and Ding, Y. 2020. "Mining Semantic Soft Factors for Credit Risk Evaluation in Peer-to-Peer Lending," *Journal of Management Information Systems*, 37(1), pp. 282-308.

Wolf, T., Chaumond, J., Debut, L., Sanh, V., Delangue, C., Moi, A., Cistac, P., Funtowicz, M., Davison, J., Shleifer, S., and Louf, R. 2020. "Transformers: State-of-the-Art Natural Language Processing," *Proceedings of EMNLP 2020: Demonstrations*, pp. 38-45.

Wu, J., Huang, L., and Zhao, J. L. 2019. "Operationalizing Regulatory Focus in the Digital Age: Evidence from an E-Commerce Context," *MIS Quarterly*, 43(3), pp. 745-764.

Yang, M., Adomavicius, G., Burtch, G., and Ren, Y. 2018. "Mind the Gap: Accounting for Measurement Error and Misclassification in Variables Generated via Data Mining," *Information Systems Research*, 29(1), pp. 4-24.

Yang, Z., Dai, Z., Yang, Y., Carbonell, J., Salakhutdinov, R.R., and Le, Q.V. 2019. "XLNet: Generalized Autoregressive Pretraining for Language Understanding," *Advances in Neural Information Processing Systems*.

Yoo, E., Gu, B., and Rabinovich, E. 2019. "Diffusion on Social Media Platforms: A Point Process Model for Interaction among Similar Content,"

*Journal of Management Information Systems*, 36(4), pp. 1105-1141.

Yue, W. T., Wang, Q.-H., and Hui, K.-L. 2019. "See no evil, hear no evil? Dissecting the impact of online hacker forums," *MIS Quarterly*, 43(1), pp. 73-95.

Zhang, D., Zhou, L., Kehoe, J. L., and Kilic, I. Y. et al. 2016. "What online reviewer behaviors really matter? Effects of verbal and nonverbal behaviors on detection of fake online reviews," *Journal of Management Information Systems*, 33(2), pp. 456-481.

Zhang, K., Bhattacharyya, S., and Ram, S. 2016. "Large-Scale Network Analysis for Online Social Brand Advertising," *MIS Quarterly*, 40(4), pp. 849-868.

Zhang, W, and Ram, S. 2020. "A Comprehensive Analysis of Triggers and Risk Factors for Asthma Based on Machine Learning and Large Heterogeneous Data Sources," *MIS Quarterly*, 44(1), pp. 305-339.

Zhou, S, Qiao, Z., Du, Q., Wang, G.A., Fan, W. and Yan, X. 2018. "Measuring Customer Agility From Online Reviews Using Big Data Text Analytics." *Journal of Management Information Systems*, 35(2), pp. 510-539.